%% file: top_arxv.tex
\documentclass[final,round,authoryear,sort,comma]{elsarticle}

\usepackage[utf8]{inputenc}
\usepackage{amsmath,amsfonts,amssymb}
\usepackage{tikz}
\usepackage{graphicx}
\usepackage{comment}
\usepackage{tabularx}
\usepackage{booktabs}
\usepackage{relsize}
\usetikzlibrary{calc}
\usetikzlibrary{positioning}
\usetikzlibrary{3d}
\usetikzlibrary{plotmarks}
\usetikzlibrary{patterns}
\usetikzlibrary{backgrounds}
\usetikzlibrary{fit}
\usepackage{pgfplots}
\usepackage{siunitx}
\usepackage{color,colortbl}
\usepackage{calc}
\usepackage{relsize}

\input{defs.tex}

    \makeatletter
    \def\ps@pprintTitle{%
       \let\@oddhead\@empty
       \let\@evenhead\@empty
       \def\@oddfoot{\reset@font\hfil\thepage\hfil}
       \let\@evenfoot\@oddfoot
    }
    \makeatother

\begin{document}

\begin{frontmatter}

%\title{Reducing the Annotation Effort for 3D Deep Delineation}% by Training on 2D Annotations}
%\title{Training 3D Deep Delineation on 2D Annotations to Reduce the Labeling Workload}% by Training on 2D Annotations}
\title{Tracing in 2D to Reduce the Annotation Effort for 3D Deep Delineation}

\author{ Mateusz Kozi\'nski\fnref{mka}}
\author{ Agata Mosinska\fnref{ama}}
\author{ Mathieu Salzmann}
\author{ Pascal Fua}
\address{Computer Vision Laboratory, \'Ecole Polytechnique F\'ed\'erale de Lausanne, \\ BC309, Station 15, CH-1015 Lausanne, Switzerland}%
\fntext[mka]{M.\ Kozi\'nski was supported by the FastProof ERC Proof of Concept Grant}%
\fntext[ama]{A.\ Mosinska was supported by the Swiss National Science Foundation}%

\input{abstract.tex}

\end{frontmatter}

\input{introduction.tex}

\input{related.tex}

\input{method.tex}

\input{experiments.tex}

\input{conclusion.tex}
\section*{References}
\bibliographystyle{plainnat}
\bibliography{short,biomed,vision,learning,optim}

\end{document}

%% file: defs.tex
\newcommand{\CE}{\mathit{C}}

\newcommand{\Loss}{L}
\newcommand{\Xindex}{i}
\newcommand{\Yindex}{j}
\newcommand{\Zindex}{k}
\newcommand{\Xaxis}{{\mathrm{\Xindex}}}
\newcommand{\Yaxis}{{\mathrm{\Yindex}}}
\newcommand{\Zaxis}{{\mathrm{\Zindex}}}
\newcommand{\inds}{{\Xindex\Yindex\Zindex}}
\newcommand{\indsk}{{\Xindex,\Yindex,\Zindex}}
\newcommand{\inpel}{x}
\newcommand{\inp}{\mathbf{\inpel}}
\newcommand{\outpel}{y}
\newcommand{\outp}{\mathbf{\outpel}}
\newcommand{\foreground}{1} %\mathit{foreground}}
\newcommand{\background}{0} %\mathit{background}}
\newcommand{\ignored}{\varnothing} 
\newcommand{\hullel}{h}
\newcommand{\hull}{\mathbf{\hullel}}

\newcommand{\gt}{\tilde{\outp}}
\newcommand{\gtel}{\tilde{\outpel}}
\newcommand{\gtx}{\gt^{\Xaxis}}
\newcommand{\gty}{\gt^{\Yaxis}}
\newcommand{\gtz}{\gt^{\Zaxis}}

\newcommand{\net}{f}
\newcommand{\param}{w}

\newcommand{\twoD}{\mathrm{2D}}
\newcommand{\threeD}{\mathrm{3D}}

\newcommand{\TS}{T}
\DeclareMathOperator*{\argmin}{arg\,min}

\newcommand{\mycomment}[1]{}

\graphicspath{{fig/}}

%% file: abstract.tex
% !TEX root = top.tex
% !TEX spellcheck = en-US

\begin{abstract}

The difficulty of obtaining annotations to build training databases still slows down the adoption of recent deep learning approaches for biomedical image analysis. In this paper, we show that we can train a Deep Net to perform 3D volumetric delineation given {\em only} 2D annotations in Maximum Intensity Projections (MIP). As a consequence, we can decrease the amount of time spent annotating by a factor of two while maintaining similar performance.

Our approach is inspired by {\em space carving}, a classical technique of reconstructing complex 3D shapes from arbitrarily-positioned cameras.  We will demonstrate its effectiveness on 3D light microscopy images of neurons and retinal blood vessels and on Magnetic Resonance Angiography (MRA) brain scans.

\end{abstract}

%% file: introduction.tex
% !TEX root = top.tex
% !TEX spellcheck = en-US

\section{Introduction}

Linear structures such as blood vessels, bronchi and dendritic trees are pervasive in medical imagery. Automatically recovering their topology has therefore become critically important to fully exploit the vast amounts of data that modern imaging devices can now produce. Machine Leaning based techniques have demonstrated their effectiveness for this purpose, but usually require substantial amounts of annotated training data to reach their full potential.

Unfortunately,  annotating complex topologies in 3D volumes by means of an inherently 2D computer interface is slow and tedious. The annotator must frequently rotate and move the volume to verify the correct placement of control points and to reveal occluded details. Not only is this inherently slow, but such interactions require continuously re-displaying large amounts of data, which often exceeds the capacity of a workstation, thus introducing further delays. 

\input{workflow_1}

In this paper, we show that we can train a Deep Net to perform 3D volumetric delineation given {\em only} 2D annotations in Maximum Intensity Projections (MIP), such as those shown on the right side of Fig.~\ref{fig:workflow}. This is a major time-saver because delineating linear structures in 2D images is much easier than in 3D volumes and involves none of the difficulties mentioned above. Furthermore, semi-automated annotation tools work more smoothly on 2D than on 3D data. In short,  limiting the annotation effort to the projections leads to a considerable labor saving without compromising the performance of the trained network.

More specifically, we introduce a loss function that penalizes discrepancies between the maximum intensity projection of the predictions and the 2D annotations. We show that it yields a network that performs as well as if it had been trained using full 3D annotations. The loss is inspired by {\em space carving}, a classical approach to  reconstructing complex 3D shapes from arbitrarily-positioned cameras~\cite{Kutulakos00}. Space carving exploits the fact that visual rays corresponding to background pixels in 2D images cannot cross any foreground voxel when passing through the volume. Conversely, rays emanating from foreground pixels  have to cross at least one foreground voxel. In our case, the rays are parallel to the projection axes. The network is trained to minimize the cross-entropy between the 2D annotations and the maximum values along the rays.

Our contribution is therefore a principled approach to reducing the annotators' burden when training a Deep Net by enabling them to trace in 2D instead of 3D, while still capturing the full 3D topology of complex linear structures. We demonstrate this on 3D light microscopy images of neurons and retinal blood vessels and on Magnetic Resonance Angiography (MRA) brain scans. An earlier version of this approach first appeared in~\cite{Kozinski18}. We present here an extended version that includes a user study that demonstrates the effectiveness of our approach, as compared to more traditional ones.

%% file: workflow_1.tex
% !TEX root = top.tex
% !TEX spellcheck = en-US

\begin{figure*}
\center
\input{workflow_diag_1}
\caption{
Training a neural network to delineate 3D structures using 3D (a) and 2D (b) annotations. (a) The standard approach is to manually or semi-automatically delineate structures in 3D volumes to create ground-truth data, which can then be used to train a deep network. (b) Ours is to delineate in 2D in 2 or 3 Maximum Intensity Projections, which is easier and faster. The projections are used to compute a loss function that exploits these 2D annotations. We use it to train the network, and achieve similar performance with half as much human intervention. 
\label{fig:workflow}
}
\end{figure*}

%% file: workflow_diag_1.tex
\setlength{\tabcolsep}{0pt}
\begin{tabular}{c}
\begin{tikzpicture}
[
node style/.style={
 font=\scriptsize,
 align=center,
 fill=black!10,
 minimum height=0.8cm,
 rounded corners=.1cm
},
loss style/.style={
 font=\scriptsize,
 align=center,
 minimum height=0.8cm
},
lbl style/.style={
  font=\scriptsize,
  align=center
}
]
  \pgfmathsetmacro{\mul}{1.0}
  \pgfmathsetmacro{\w}{\mul*.125}
  \pgfmathsetmacro{\sz}{\mul*0.5}
  \node (input) [] at (0,0) {\includegraphics[width=\w\textwidth]{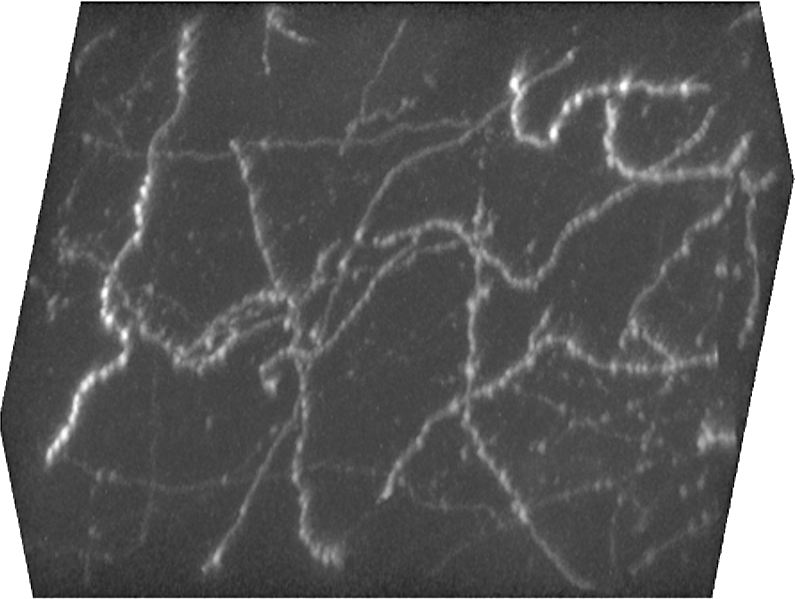}};
  \node (il) [lbl style] at (-0.1*\mul,-0.9*\mul) {3D input volume};
  \node (net) [text width=1.1*\mul cm,align=center,node style] at (2.0*\mul,0) {Neural Network};
  \node (output) [] at (4.0*\mul,0) {\includegraphics[width=\w\textwidth]{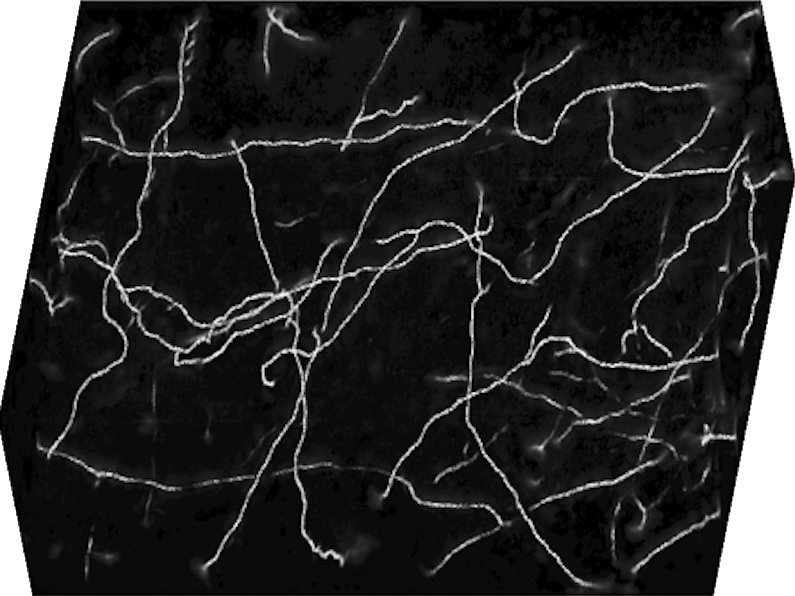}};
  \node (ol) [lbl style] at (3.9*\mul,-0.9*\mul) {3D output volume};
  \node (target) [] at (0,2.0*\mul) {\includegraphics[width=\w\textwidth]{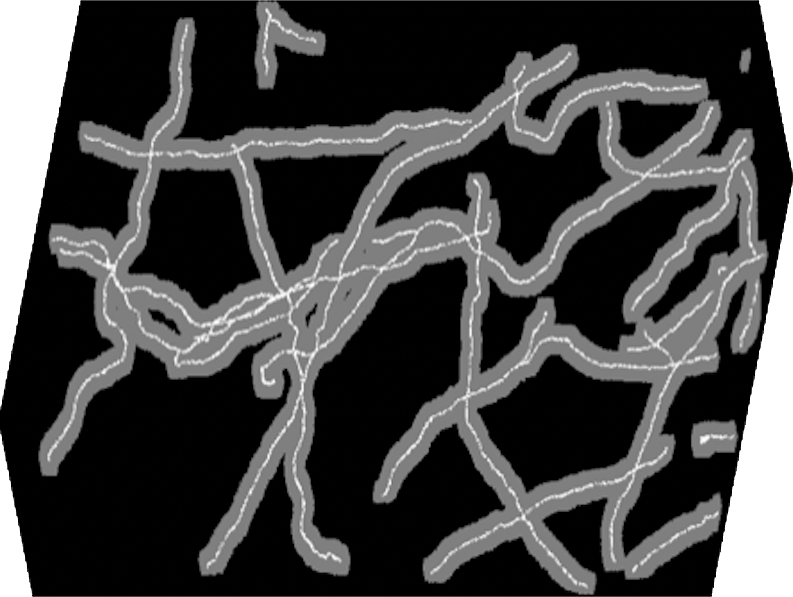}};
  \node (tl) [lbl style] at (-0.1*\mul,1.15*\mul) {3D annotations};
  \node (ce) [text width=1.0*\mul cm,align=center,node style] at (6.0*\mul,0) {Cross Entropy};
  \node (loss) [text width=0.5*\mul cm,align=center,loss style] at (7.3*\mul,0) {Loss};
  \draw[-latex] (input) -- (net);
  \draw[-latex] (net) -- (output);
  \draw[-latex] (output) -- (ce);
  \draw[-latex] (ce) -- (loss);
  \draw[-latex,rounded corners=0.025cm] (target) -| (ce);

  \node (costly) [] at (2.2*\mul,2.5*\mul) {\textcolor{red}{Costly!}};
  \draw[-latex,ultra thick,red](costly) -- (target);
\end{tikzpicture}
\\
\footnotesize (a) Standard training procedure.
\\
\\  
\begin{tikzpicture}
[
node style/.style={
 font=\scriptsize,
 align=center,
 fill=black!10,
 minimum height=0.8cm,
 rounded corners=.1cm
},
loss style/.style={
 font=\scriptsize,
 align=center,
 minimum height=0.8cm
},
lbl style/.style={
  font=\scriptsize,
  align=center
}
]
  \pgfmathsetmacro{\mul}{1.0}
  \pgfmathsetmacro{\w}{\mul*.125}
  \pgfmathsetmacro{\sz}{\mul*0.5}
  \node (input) [] at (0,0) {\includegraphics[width=\w\textwidth]{MIP_input.png}};
  \node (il) [lbl style] at (-0.1*\mul,-0.9*\mul) {3D input volume};
  \node (net) [text width=1.1*\mul cm,align=center,node style] at (1.8*\mul,0) {Neural Network};
  \node (output) [] at (3.6*\mul,0) {\includegraphics[width=\w\textwidth]{MIP_prediction.png}};
  \node (ol) [lbl style] at (3.4*\mul,-0.9*\mul) {3D output volume};
  \node (target) [] at (0,2.0*\mul) {\includegraphics[width=\w\textwidth]{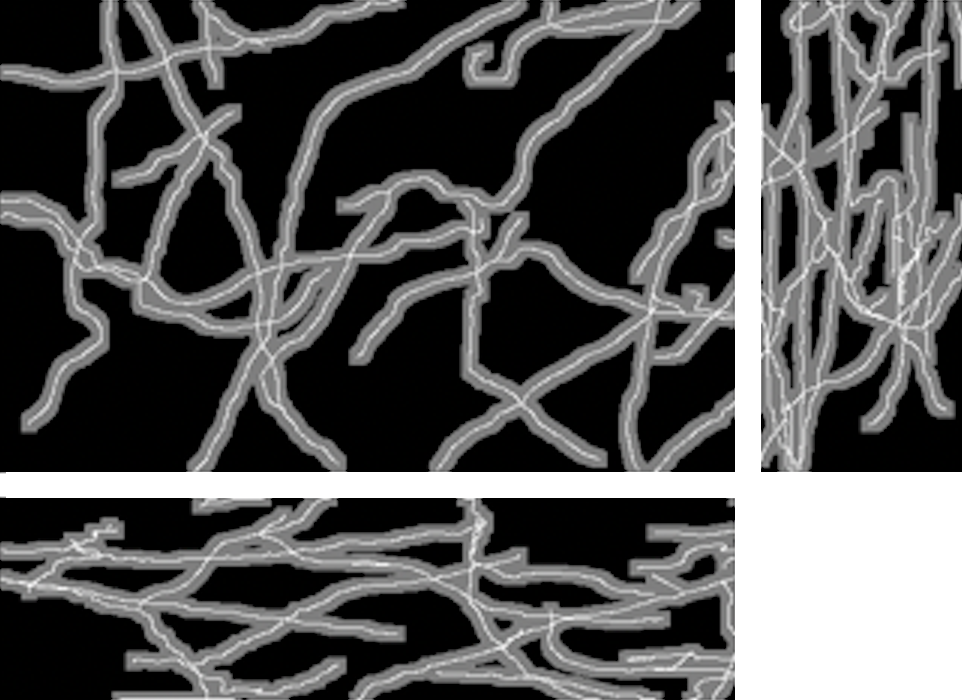}};
  \node (tl) [lbl style] at (-0.1*\mul,1.15*\mul) {2D annotations};
  \node (project) [text width=1.0*\mul cm,align=center,node style] at (5.4*\mul,0) {Project};
  \node (ce) [text width=1.0*\mul cm,align=center,node style] at (9.0*\mul,0) {Cross Entropy};
  \node (loss) [text width=0.5*\mul cm,align=center,loss style] at (10.3*\mul,0) {Loss};
  \node (proj_output) [] at (7.2*\mul,0) {\includegraphics[width=\w\textwidth]{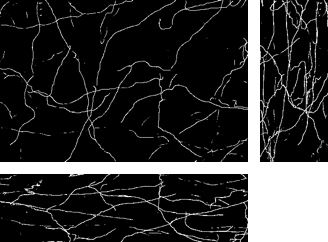}};
  \node (pl) [lbl style] at (7.2*\mul,-0.9*\mul) {2D output projection};
  \draw[-latex] (input) -- (net);
  \draw[-latex] (net) -- (output);
  \draw[-latex] (output) -- (project);
  \draw[-latex] (ce) -- (loss);
  \draw[-latex] (project) -- (proj_output);
  \draw[-latex] (proj_output) -- (ce);
  \draw[-latex,rounded corners=0.025cm] (target) -| (ce);
  \draw[-,blue!80,thick,rounded corners=0.05cm] (4.6*\mul,-1.1*\mul) -- ++(5.1*\mul,0) -- ++(0,1.9*\mul) -- ++(-5.1*\mul,0) -- cycle;
  \node (ll) [lbl style,anchor=west,blue!80] at (4.6*\mul,1.0*\mul) {The proposed loss};

  \node (costly) [] at (2.6*\mul,2.5*\mul) {\textcolor{green!50!black}{Far less costly!}};
  \draw[-latex,ultra thick,green!50!black](costly) -- (target);
\end{tikzpicture}
\\
\footnotesize (b) Our training procedure.
\\
\end{tabular}

%% file: related.tex
% !TEX root = top.tex
% !TEX spellcheck = en-US

\section{Related Work}
\label{sec:related}

Delineation is a broad research topic. It operates on structures as different as roads~\citep{Mnih10,Mnih13,Wegner13,Mattyus17}, blood vessels~\citep{Ganin14,Maninis16}, bronchi~\citep{Meng17}, neurites~\citep{Sironi16a,Peng17}, and cell membranes~\citep{Mosinska17}, imaged using many different modalities. In this paper, we specifically address 3D delineation where the input is a volume, as opposed to a collection of ordered, but unregistered slices~\citep{Funke12}.

Early approaches to delineation of 3D curvilinear structures relied on filters manually designed to respond strongly to tubular segments~\citep{Sato98,Frangi98,Law08,Turetken13c}. They do not require to be trained, but their performance degrades when the structures become irregular and the images noisy. This has led to the emergence of machine learning-based methods that can cope with such difficulties, given enough annotated data~\citep{Becker13b,Breitenreicher13,Sironi16a,Peng17,Meng17}. The most recent ones of these~\citep{Peng17,Meng17} rely on a combination of Deep Learning and adaptive exploration of the light microscopy images, and Computed Tomography (CT) scans. 

However, using Machine Learning, and Deep Learning in particular, requires large amounts of annotated training data. Furthermore, annotating 3D stacks is much more labor-intensive than annotating 2D images. Only true experts, whose time is precious, are able to orient themselves and follow complex structures in large volumes~\citep{Peng14}. Until now, this problem has been handled by developing better ways to visualize and interact with image stacks~\citep{Vitanovski09,Peng17}. \cite{Cicek16} annotated only a few slices of a volume and computed the loss using only them. The technique of~\cite{Peng14}, like ours, allows the annotator to trace a linear structure in a maximum intensity projection and then attempts to guess the value of the third coordinate using a simple heuristic. While effective when the structures are relatively sparse, this can easily get confused as the scene becomes more cluttered. 

The originality of our approach is to introduce a method that relies solely on 2D annotations in Maximum Intensity Projections, yet captures the 3D structure of complex linear structures when the projections are used jointly.

%% file: method.tex
% !TEX root = main.tex
% !TEX spellcheck = en-US

\section{Method}
\label{sec:method}

\subsection{From 3D to 2D Annotations}
\label{sec:2DAnnotations}

Let us first consider the problem of training a neural network $\net_\param$, parameterized by weights $\param$, to segment linear structures within 3D image stacks, given a training set $\TS$ of 
pairs $(\inp,\gt)$, where each 3D image $\inp$ is accompanied by the corresponding volumetric ground-truth annotations $\gt$.
We denote the elements of $\inp$ and $\gt$ by $\inpel_\inds$ and $\gtel_\inds$, where $\indsk$ index the positions of the elements within the volumes. The ground-truth labels take a value in the set $\{\foreground,\background,\ignored\}$, which indicate the presence of a linear structure in voxel $\indsk$ if $\gtel_\inds=\foreground$, the absence of a linear structure if $\gtel_\inds=\background$, and uncertainty of the annotator if $\gtel_\inds=\ignored$. Delineation can then be cast as a binary segmentation problem by simply ignoring the voxels labeled as $\ignored$ during training. The network output $\outp=\net_\param(\inp)$ has the same size as the input and contains probabilities of presence of a linear structure in each voxel. To train the network, we find
\begin{equation}
\argmin_{w}  \sum_{(\inp,\gt)\in\TS} \sum_{\indsk} \Loss(\net_\param(\inp)_\inds,\gtel_\inds) \, ,
	 \label{eq:loss3D}
\end{equation}
where $\net_\param()_\inds$ denotes voxel $\indsk$ of the prediction, and the loss $\Loss(\outpel,\gtel)$ is taken to be the cross entropy
%\begin{equation}
$\CE(\outpel,\gtel)=[\gtel=\foreground] \log \outpel+[\gtel=\background] \log(1-\outpel)$,
%\end{equation}
where $[\cdot]$ is the Iverson bracket.
As discussed in the introduction, the drawback of this approach is that generating the ground-truth labels $\gt$ in sufficient numbers to train a deep network is tedious and expensive when operating on large volumes.

To alleviate this problem, we reformulate the loss function of Eq.~\ref{eq:loss3D} so that it can exploit annotated Maximum Intensity Projections (MIPs) of the input volumes. A MIP of volume $\inp$ along direction $\Xaxis$, which we denote as $\inp^\Xaxis$, is a 2D image with elements $\inpel^\Xaxis_{\Yindex\Zindex}=\max_{\Xindex} \inpel_\inds$. Annotating MIPs is easy when the structures of interest have high intensity and are clearly visible in the projections. A MIP annotation $\gt^\Xaxis$ of the projection $\inp^\Xaxis$ is a 2D image with elements $\gtel^\Xaxis_{\Yindex\Zindex}\in\{\foreground,\background,\ignored\}$, where the labels have the same interpretation as the ones used for annotating in 3D.
%which can also be thought of as $\gtel^\Xaxis_{\Yindex\Zindex}=\max_{\Xindex} \gtel_\inds$. 
MIPs of the volume along the directions $\Yaxis$ and $\Zaxis$, and their annotations, are defined similarly. 

The key property of MIP annotations, is that $\gtel^\Xaxis_{\Yindex\Zindex}=\background$ tells us that {\em all} voxels of the input column $\Yindex\Zindex$ contain background. 
To see that the property really holds, let us assume an idealized case where the Maximum Intensity Projection operation, and the act of annotation, preserve the linear structures. In other words, we assume that, if the training volume contains an image of a linear structure in any voxel of column $\Yindex\Zindex$, then this linear structure will necessarily be visible in the Maximum Intensity Projection, in pixel $\inp^\Xaxis_{\Yindex\Zindex}$, and will be annotated as foreground in the MIP annotation, so that $\gtel^\Xaxis_{\Yindex\Zindex}=\foreground$.
Under these assumptions, by De Morgan's law, $\gtel^\Xaxis_{\Yindex\Zindex}\ne\foreground$ implies that no voxel of the column $\Yindex\Zindex$ is of foreground class.

It is exactly this property that enables establishing a link between training on MIP annotations and space carving. In space carving, a single background pixels of an image of a 3D scene is used to classify many voxels of scene reconstruction as background, effectively carving out the reconstructed shape. When training a network on MIP annotations, a pixel annotated as background could be used to constrain many voxels of the prediction to belong to the background class, thus generating an error signal for these voxels. In practice, instead of enforcing this constraint directly, we formulate a loss function that capitalizes on this observation implicitly. 
To that end, we define the max-projection $\net^\Xaxis_\param(\inp)$ along direction $\Xaxis$ of the network output as the image with elements $\net^\Xaxis_\param(\inp)_{\Yindex\Zindex}=\max_{\Xindex} \net_\param(\inp)_\inds$. We proceed similarly for directions $\Yaxis$ and $\Zaxis$. We then define the loss as
%
% \argmin_{w} 
\begin{multline}
   \sum_{(\inp,\gt)\in\TS}  \Big( 
     \sum_{\Yindex\Zindex} \Loss\big(\net^\Xaxis_\param(\inp)_{\Yindex\Zindex},\gtel^{\Xaxis}_{\Yindex\Zindex}\big) 
    +\sum_{\Xindex\Zindex} \Loss\big(\net^\Yaxis_\param(\inp)_{\Xindex\Zindex},\gtel^{\Yaxis}_{\Xindex\Zindex}\big) \\
    +\sum_{\Xindex\Yindex} \Loss\big(\net^\Zaxis_\param(\inp)_{\Xindex\Yindex},\gtel^{\Zaxis}_{\Xindex\Yindex}\big) 
  \Big)\, .
  \label{eq:loss2D}
\end{multline}
To see the analogy to space carving, note that, by its definition, $\net^\Xaxis_\param(\inp)_{\Yindex\Zindex}$ upper bounds the predicted probability of presence of a linear structure in column $\Yindex\Zindex$. Eq.~\ref{eq:loss2D} penalizes large values of this upper bound whenever $\gtel^\Xaxis_{\Yindex\Zindex}=\background$. In other words, a single background label in a 2D annotation results in minimization of a whole column of predictions, mimicking space carving. When $\gtel^\Xaxis_{\Yindex\Zindex}=\foreground$, minimizing the loss increases the largest prediction in the column. The latter one might be placed off a linear structure, but it is then likely to be penalized by a component of the loss defined for another projection. 

Another interesting observation is that the loss~\eqref{eq:loss2D} yields very sparse gradients. Indeed, each projection gives rise to just a single nonzero element for a whole column of the gradient. At a first glance, this threatens to compromise the performance of the trained network. 
However, the nonzero elements are not distributed randomly over the gradient tensor. As demonstrated in sections~\ref{sssec:performance} and~\ref{sec:3Dvs2D}, in practice the networks trained with 2D annotations perform on par with ones trained on the full 3D annotations, and the space carving mechanism seems to be the secret behind this surprising result. 

\subsection{Visual Hull for Training on Cropped Volumes}
\label{sec:spaceCarving}
\input{trainingOnCrops}

Due to memory limitations, the annotated training volumes are typically cropped into sub-volumes and the MIP annotations can be cropped to match. However, the cropped annotations may then contain labels for structures located outside the volume crop, as illustrated  by Fig.~\ref{fig:trainingOnCrops}. To reduce the influence of these extraneous annotations, we use another element of the space carving theory, the visual hull $\hull$. $\hull$ is a volume containing the original one, and constructed from its projections~\citep{Kutulakos00}. A toy example of a visual hull created from 2D projections of a volume is presented in Fig.~\ref{fig:filter}(a). We define it more precisely below.
\input{hullCrop}

We first introduce the definition of the hull for the classic, binary case. Given three orthogonal MIP annotations $\gtx$, $\gty$, $\gtz$, with elements $\gtel^\Xaxis_{\Yindex\Zindex},\gtel^\Yaxis_{\Zindex\Zindex},\gtel^\Zaxis_{\Xindex\Yindex}\in\{\background,\foreground\}$, we define the hull $\hull$ as a binary volume with elements 
\begin{equation}
\hullel_\inds=  
\begin{cases}
\foreground & \text{if }  \gtel^\Xaxis_{\Yindex\Zindex} = \foreground \wedge \gtel^\Yaxis_{\Xindex\Zindex} = \foreground \wedge \gtel^\Zaxis_{\Xindex\Yindex} = \foreground ,\\
\background & \text{otherwise.}
\end{cases}
\end{equation}
By construction, an element of the hull $\hullel_\inds=\foreground$ if and only if \emph{all} of its projections are labeled as foreground. In our context, a foreground voxel outside a crop only produces an incorrect label in \emph{a single} projection, as demonstrated in Fig.~\ref{fig:trainingOnCrops}. As shown in Fig.~\ref{fig:filter}(b), we can eliminate such false positive labels by projecting the visual hull back to the 2D annotations and discarding the labels that fall outside of the projection of the visual hull. However, this technique fails to eliminate these false positive labels, for which in each of the remaining projection annotations another positive label exists with the same coordinate along the common dimension. Such situation is illustrated in Fig.~\ref{fig:filter}(c). Our experiments show that such rare events have little impact on the performance of the trained network.

As stated in section \ref{sec:2DAnnotations}, in practice our annotations are defined in terms of a ternary set of labels, with the additional label $\ignored$, allowing the annotator to skip labeling a pixel if he is not certain of its class. In our experiments, we also use this additional label to create margins around thin annotations of centerlines of linear structures, in order to account for the ambiguity in defining the latter.
In order to apply the visual-hull-based technique to eliminate false positive labels from such ternary MIP annotations, we reduce the number of classes in the annotations to two when constructing the visual hull. More precisely, we consider the foreground label and the label encoding the uncertainty of the annotator as positive, and the background label as negative. Then, for each projection annotation, we project the hull along the same direction and suppress all the positive and uncertainty labels that collide with the negative class in the projection of the hull. In other words, we propagate the background labels between projections via the visual hull.

In the experiments presented in section \ref{sec:exp} we train a deep network on 3, 2 or 1 MIP annotations per volume. The definition of the visual hull presented above trivially generalizes to the 2-MIP cases, and the procedure is not performed when only 1 MIP annotation is used.

\subsection{Implementation}
\label{sec:impl}

In practice, we implemented $\net_\param$ as a U-Net style network~\citep{Ronneberger15}. Specifically, we made the original convolution-ReLU blocks residual, and only used two max-pooling operations instead of the usual four, which resulted in a more compact network that fits in memory even with larger volume crops.
In all our experiments, we trained the network for 200K iterations, using the ADAM update scheme~\citep{Kingma15} with momentum of $0.9$, weight decay $10^{-4}$ and step size $10^{-5}$.

%% file: trainingOnCrops.tex
\begin{figure}
\centering
\begin{tikzpicture}
\node[anchor=south west,inner sep=0] (image) at (0,0,0) {\includegraphics[width=0.3\columnwidth]{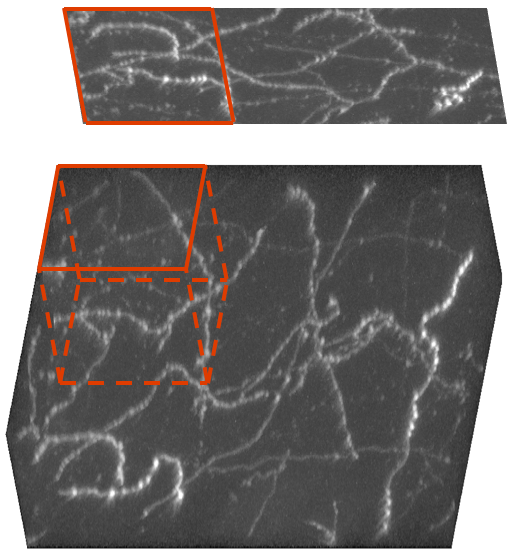}};
\draw[latex-,ultra thick,green] (1.3,3.6,0) -- ++(0.4,0.0,0.0);
\draw[latex-,ultra thick,green] (0.91,0.65,0) -- ++(0.4,0.0,0.0);
\node[rotate=90,anchor=south] at (0,1.5,0) (a) {\footnotesize original volume};
\node[rotate=90,anchor=south] at (0,3.5,0) (b) {\footnotesize MIP};
\end{tikzpicture}
\caption{
When training on MIP annotations, using volume crops (brown cube) may lead to situations where, a crop of a MIP annotation (brown rectangle) contains labels of linear structures from outside of the volume crop (marked with green arrows). This annotation noise could adversely influence performance of the trained network.
\label{fig:trainingOnCrops}
}
\end{figure}
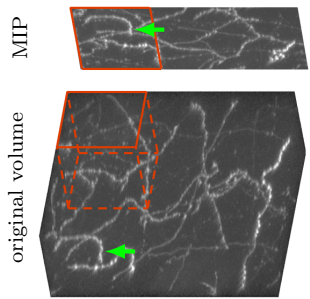

%% file: hullCrop.tex
% !TEX root = ../main.tex
% !TEX spellcheck = en-US

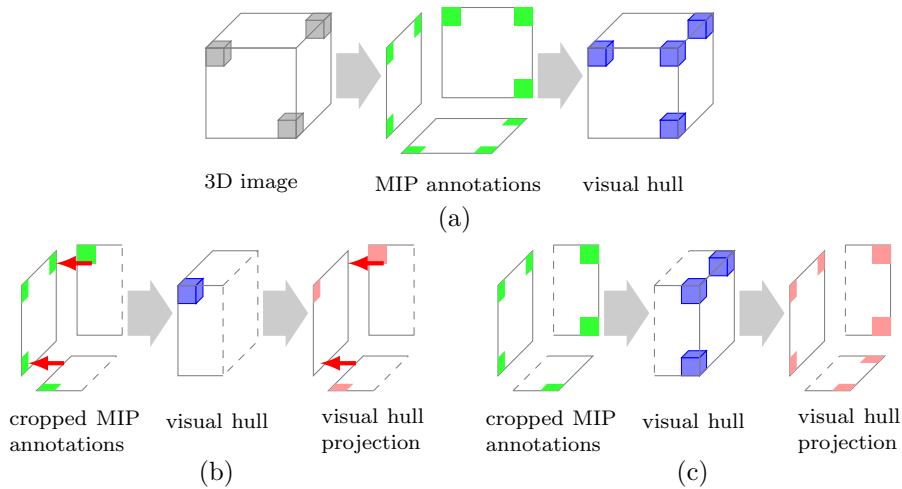
\begin{figure*}
\input{hull_diagram.tex}
\caption{
Handling cropped volumes. \emph{(a)} A 3D volume with three foreground voxels, the annotations of its MIPs in green, and the visual hull computed from these in blue.  
\emph{(b)} The volume has been cropped so that only the left half remains. The annotations have been cropped to match, leaving a single blue voxel in the visual hull. Reprojecting it into the MIPs lets us eliminate the extraneous annotations, indicated with red arrows. 
\emph{(c)}  However, there are situations such as the one depicted here, where some will survive.}
\label{fig:filter}
\end{figure*}

%% file: hull_diagram.tex
% !TEX root = ../main.tex
% !TEX spellcheck = en-US

\center
\setlength{\tabcolsep}{0pt}
\begin{tabular}{c@{\hspace{0.5cm}}c}
\multicolumn{2}{c}{
{
\begin{tikzpicture}[scale=0.8]
\pgfmathsetmacro{\cubesize}{1.5}
\pgfmathsetmacro{\cubeoffset}{0.75}
\pgfmathsetmacro{\rad}{0.3}
\pgfmathsetmacro{\projoffset}{0.25}
\node[] at (0,-\cubeoffset-\projoffset-0.5,\cubeoffset) (b) {\footnotesize 3D image};

\filldraw[fill=gray!50,draw=gray] (\cubeoffset-\cubesize+\rad,\cubeoffset,\cubeoffset) -- ++(-\rad,0,0) -- ++(0,-\rad,0) -- ++(\rad,0,0) -- cycle;
\filldraw[fill=gray!50,draw=gray] (\cubeoffset-\cubesize+\rad,\cubeoffset,\cubeoffset) -- ++(-\rad,0,0) -- ++(0,0,-\rad) -- ++(\rad,0,0) -- cycle;
\filldraw[fill=gray!50,draw=gray] (\cubeoffset-\cubesize+\rad,\cubeoffset,\cubeoffset) -- ++(0,0,-\rad) -- ++(0,-\rad,0) -- ++(0,0,\rad) -- cycle;
\filldraw[fill=gray!50,draw=gray] (\cubeoffset,\cubeoffset-\cubesize+\rad,\cubeoffset) -- ++(-\rad,0,0) -- ++(0,-\rad,0) -- ++(\rad,0,0) -- cycle;
\filldraw[fill=gray!50,draw=gray] (\cubeoffset,\cubeoffset-\cubesize+\rad,\cubeoffset) -- ++(-\rad,0,0) -- ++(0,0,-\rad) -- ++(\rad,0,0) -- cycle;
\filldraw[fill=gray!50,draw=gray] (\cubeoffset,\cubeoffset-\cubesize+\rad,\cubeoffset) -- ++(0,0,-\rad) -- ++(0,-\rad,0) -- ++(0,0,\rad) -- cycle;
\filldraw[fill=gray!50,draw=gray] (\cubeoffset,\cubeoffset,\cubeoffset-\cubesize+\rad) -- ++(-\rad,0,0) -- ++(0,-\rad,0) -- ++(\rad,0,0) -- cycle;
\filldraw[fill=gray!50,draw=gray] (\cubeoffset,\cubeoffset,\cubeoffset-\cubesize+\rad) -- ++(-\rad,0,0) -- ++(0,0,-\rad) -- ++(\rad,0,0) -- cycle;
\filldraw[fill=gray!50,draw=gray] (\cubeoffset,\cubeoffset,\cubeoffset-\cubesize+\rad) -- ++(0,0,-\rad) -- ++(0,-\rad,0) -- ++(0,0,\rad) -- cycle;
%\draw[gray!80,dashed] (\cubeoffset-\rad,\cubeoffset-\rad,\cubeoffset-\rad) circle [radius=\rad];
\draw[gray] (\cubeoffset,\cubeoffset,\cubeoffset) -- ++(-\cubesize,0,0) -- ++(0,-\cubesize,0) -- ++(\cubesize,0,0); % -- cycle;
\draw[gray] (\cubeoffset,\cubeoffset,\cubeoffset) -- ++(-\cubesize,0,0) -- ++(0,0,-\cubesize) -- ++(\cubesize,0,0); % -- cycle;
\draw[gray] (\cubeoffset,\cubeoffset,\cubeoffset) -- ++(0,0,-\cubesize) -- ++(0,-\cubesize,0) -- ++(0,0,\cubesize) -- cycle;
%\draw[gray,dashed] (\cubeoffset,\cubeoffset,\cubeoffset) ++(-\cubesize,0,-\cubesize) -- ++(0,-\cubesize,0) -- ++(\cubesize,0,0) ++(-\cubesize,0,0) -- ++(0,0,\cubesize);
\pgfmathsetmacro{\arrowoffsetx}{1.5*\cubeoffset}
\pgfmathsetmacro{\arrowoffsety}{0.5*\cubeoffset}
\pgfmathsetmacro{\arrowoffsetz}{0}
\pgfmathsetmacro{\projoffset}{.25}
\pgfmathsetmacro{\xcubeoffset}{4.0}
\fill[black!20] (\arrowoffsetx,\arrowoffsety,\arrowoffsetz) -- ++(.25*\cubesize,0,0) -- ++(0,.125*\cubesize,0) -- ++(.25*\cubesize,-\arrowoffsety-.125*\cubesize,0) -- ++(-.25*\cubesize,-\arrowoffsety-.125*\cubesize,0) -- ++(0,.125*\cubesize,0) -- ++(-.25*\cubesize,0,0) -- cycle;
\draw[gray] (\xcubeoffset,\cubeoffset,\cubeoffset) ++(-\cubesize-\projoffset,0,0) -- ++(0,0,-\cubesize) -- ++(0,-\cubesize,0) -- ++(0,0,\cubesize) -- cycle;
\draw[gray] (\xcubeoffset,\cubeoffset,\cubeoffset) ++(0,-\cubesize-\projoffset,0) ++(0,0,-\cubesize) -- ++(-\cubesize,0,0) -- ++(0,0,\cubesize) -- ++(\cubesize,0,0) -- cycle;
\draw[gray] (\xcubeoffset,\cubeoffset,\cubeoffset) ++(0,0,-\cubesize-\projoffset) -- ++(-\cubesize,0,0) -- ++(0,-\cubesize,0) -- ++(\cubesize,0,0) -- cycle;

\node[] at (\xcubeoffset-0.75*\cubeoffset,-\cubeoffset-\projoffset-0.5,\cubeoffset) (b) {\footnotesize MIP annotations};
\fill[green!80] (\xcubeoffset,\cubeoffset,\cubeoffset) ++(-\cubesize-\projoffset,0,0) -- ++(0,0,-\rad) -- ++(0,-\rad,0) -- ++(0,0,\rad) -- cycle;
\fill[green!80] (\xcubeoffset,\cubeoffset,\cubeoffset) ++(-\cubesize-\projoffset,-\cubesize+\rad,0) -- ++(0,0,-\rad) -- ++(0,-\rad,0) -- ++(0,0,\rad) -- cycle;
\fill[green!80] (\xcubeoffset,\cubeoffset,\cubeoffset) ++(-\cubesize-\projoffset,0,-\cubesize+\rad) -- ++(0,0,-\rad) -- ++(0,-\rad,0) -- ++(0,0,\rad) -- cycle;
\fill[green!80] (\xcubeoffset,\cubeoffset,\cubeoffset) ++(-\cubesize+\rad,-\cubesize-\projoffset,0) -- ++(0,0,-\rad) -- ++(-\rad,0,0) -- ++(0,0,\rad) -- cycle;
\fill[green!80] (\xcubeoffset,\cubeoffset,\cubeoffset) ++(0,-\cubesize-\projoffset,0) -- ++(0,0,-\rad) -- ++(-\rad,0,0) -- ++(0,0,\rad) -- cycle;
\fill[green!80] (\xcubeoffset,\cubeoffset,\cubeoffset) ++(0,-\cubesize-\projoffset,-\cubesize+\rad) -- ++(0,0,-\rad) -- ++(-\rad,0,0) -- ++(0,0,\rad) -- cycle;
\fill[green!80] (\xcubeoffset,\cubeoffset,\cubeoffset) ++(-\cubesize+\rad,0,-\cubesize-\projoffset) -- ++(-\rad,0,0) -- ++(0,-\rad,0) -- ++(\rad,0,0) -- cycle;
\fill[green!80] (\xcubeoffset,\cubeoffset,\cubeoffset) ++(0,0,-\cubesize-\projoffset) -- ++(-\rad,0,0) -- ++(0,-\rad,0) -- ++(\rad,0,0) -- cycle;
\fill[green!80] (\xcubeoffset,\cubeoffset,\cubeoffset) ++(0,-\cubesize+\rad,-\cubesize-\projoffset) -- ++(-\rad,0,0) -- ++(0,-\rad,0) -- ++(\rad,0,0) -- cycle;

\pgfmathsetmacro{\arrowoffsetx}{1.5*\cubeoffset+3.35}
\pgfmathsetmacro{\arrowoffsety}{0.5*\cubeoffset}
\pgfmathsetmacro{\arrowoffsetz}{0}
\pgfmathsetmacro{\projoffset}{.25}
\pgfmathsetmacro{\xcubeoffset}{7.1}
\fill[black!20] (\arrowoffsetx,\arrowoffsety,\arrowoffsetz) -- ++(.25*\cubesize,0,0) -- ++(0,.125*\cubesize,0) -- ++(.25*\cubesize,-\arrowoffsety-.125*\cubesize,0) -- ++(-.25*\cubesize,-\arrowoffsety-.125*\cubesize,0) -- ++(0,.125*\cubesize,0) -- ++(-.25*\cubesize,0,0) -- cycle;
\filldraw[fill=blue!50,draw=blue] (\xcubeoffset-\cubesize+\rad,\cubeoffset,\cubeoffset) -- ++(-\rad,0,0) -- ++(0,-\rad,0) -- ++(\rad,0,0) -- cycle;
\filldraw[fill=blue!50,draw=blue] (\xcubeoffset-\cubesize+\rad,\cubeoffset,\cubeoffset) -- ++(-\rad,0,0) -- ++(0,0,-\rad) -- ++(\rad,0,0) -- cycle;
\filldraw[fill=blue!50,draw=blue] (\xcubeoffset-\cubesize+\rad,\cubeoffset,\cubeoffset) -- ++(0,0,-\rad) -- ++(0,-\rad,0) -- ++(0,0,\rad) -- cycle;
\filldraw[fill=blue!50,draw=blue] (\xcubeoffset,\cubeoffset-\cubesize+\rad,\cubeoffset) -- ++(-\rad,0,0) -- ++(0,-\rad,0) -- ++(\rad,0,0) -- cycle;
\filldraw[fill=blue!50,draw=blue] (\xcubeoffset,\cubeoffset-\cubesize+\rad,\cubeoffset) -- ++(-\rad,0,0) -- ++(0,0,-\rad) -- ++(\rad,0,0) -- cycle;
\filldraw[fill=blue!50,draw=blue] (\xcubeoffset,\cubeoffset-\cubesize+\rad,\cubeoffset) -- ++(0,0,-\rad) -- ++(0,-\rad,0) -- ++(0,0,\rad) -- cycle;
\filldraw[fill=blue!50,draw=blue] (\xcubeoffset,\cubeoffset,\cubeoffset-\cubesize+\rad) -- ++(-\rad,0,0) -- ++(0,-\rad,0) -- ++(\rad,0,0) -- cycle;
\filldraw[fill=blue!50,draw=blue] (\xcubeoffset,\cubeoffset,\cubeoffset-\cubesize+\rad) -- ++(-\rad,0,0) -- ++(0,0,-\rad) -- ++(\rad,0,0) -- cycle;
\filldraw[fill=blue!50,draw=blue] (\xcubeoffset,\cubeoffset,\cubeoffset-\cubesize+\rad) -- ++(0,0,-\rad) -- ++(0,-\rad,0) -- ++(0,0,\rad) -- cycle;
\filldraw[fill=blue!50,draw=blue] (\xcubeoffset,\cubeoffset,\cubeoffset) -- ++(-\rad,0,0) -- ++(0,-\rad,0) -- ++(\rad,0,0) -- cycle;
\filldraw[fill=blue!50,draw=blue] (\xcubeoffset,\cubeoffset,\cubeoffset) -- ++(-\rad,0,0) -- ++(0,0,-\rad) -- ++(\rad,0,0) -- cycle;
\filldraw[fill=blue!50,draw=blue] (\xcubeoffset,\cubeoffset,\cubeoffset) -- ++(0,0,-\rad) -- ++(0,-\rad,0) -- ++(0,0,\rad) -- cycle;
%\draw[gray!80,dashed] (\cubeoffset-\rad,\cubeoffset-\rad,\cubeoffset-\rad) circle [radius=\rad];
\draw[gray] (\xcubeoffset,\cubeoffset,\cubeoffset) -- ++(-\cubesize,0,0) -- ++(0,-\cubesize,0) -- ++(\cubesize,0,0); % -- cycle;
\draw[gray] (\xcubeoffset,\cubeoffset,\cubeoffset) -- ++(-\cubesize,0,0) -- ++(0,0,-\cubesize) -- ++(\cubesize,0,0); % -- cycle;
\draw[gray] (\xcubeoffset,\cubeoffset,\cubeoffset) -- ++(0,0,-\cubesize) -- ++(0,-\cubesize,0) -- ++(0,0,\cubesize) -- cycle;
\node[] at (\xcubeoffset-0.75,-\cubeoffset-\projoffset-0.5,\cubeoffset) (c) {\footnotesize visual hull};
%\draw[gray,dashed] (\cubeoffset,\cubeoffset,\cubeoffset) ++(-\cubesize,0,-\cubesize) -- ++(0,-\cubesize,0) -- ++(\cubesize,0,0) ++(-\cubesize,0,0) -- ++(0,0,\cubesize);
%\filldraw[fill=gray!50,draw=gray] (\xcubeoffset-.5*\cubesize+\rad,\cubeoffset,\cubeoffset) -- ++(-\rad,0,0) -- ++(0,-\rad,0) -- ++(\rad,0,0) -- cycle;
%\filldraw[fill=gray!50,draw=gray] (\xcubeoffset-.5*\cubesize+\rad,\cubeoffset,\cubeoffset) -- ++(-\rad,0,0) -- ++(0,0,-\rad) -- ++(\rad,0,0) -- cycle;
%\filldraw[fill=gray!50,draw=gray] (\xcubeoffset-.5*\cubesize+\rad,\cubeoffset,\cubeoffset) -- ++(0,0,-\rad) -- ++(0,-\rad,0) -- ++(0,0,\rad) -- cycle;
%\draw[gray!80,dashed] (\cubeoffset-\rad,\cubeoffset-\rad,\cubeoffset-\rad) circle [radius=\rad];
%\draw[gray] (\xcubeoffset,\cubeoffset,\cubeoffset) -- ++(-.5*\cubesize,0,0) -- ++(0,-\cubesize,0) -- ++(.5*\cubesize,0,0); % -- cycle;
%\draw[gray] (\xcubeoffset,\cubeoffset,\cubeoffset) -- ++(-.5*\cubesize,0,0) -- ++(0,0,-\cubesize) -- ++(.5*\cubesize,0,0); % -- cycle;
%\draw[gray,dashed] (\xcubeoffset,\cubeoffset,\cubeoffset) -- ++(0,0,-\cubesize) -- ++(0,-\cubesize,0) -- ++(0,0,\cubesize) -- cycle;
\end{tikzpicture}
}
}
\\[-1mm]
\multicolumn{2}{c}{(a)} \\
%\resizebox{!}{3.5cm}{
{
\begin{tikzpicture}[scale=0.8]
\clip (-0.8,-2.5) rectangle + (7.0,3.8);
\pgfmathsetmacro{\cubesize}{1.5}
\pgfmathsetmacro{\cubeoffset}{0.75}
\pgfmathsetmacro{\rad}{0.3}
\pgfmathsetmacro{\projoffset}{.25}
\draw[gray] (\cubeoffset,\cubeoffset,\cubeoffset) ++(-0.5*\cubesize-\projoffset,0,0) -- ++(0,0,-\cubesize) -- ++(0,-\cubesize,0) -- ++(0,0,\cubesize) -- cycle;
\draw[gray] (\cubeoffset,\cubeoffset,\cubeoffset) ++(0,-\cubesize-\projoffset,0) ++(0,0,-\cubesize) -- ++(-0.5*\cubesize,0,0) -- ++(0,0,\cubesize) -- ++(0.5*\cubesize,0,0);
\draw[gray] (\cubeoffset,\cubeoffset,\cubeoffset) ++(0,0,-\cubesize-\projoffset) -- ++(-.5*\cubesize,0,0) -- ++(0,-\cubesize,0) -- ++(.5*\cubesize,0,0) ;
\draw[gray,dashed] (\cubeoffset,\cubeoffset,\cubeoffset) ++(0,-\cubesize-\projoffset,0) -- ++(0,0,-\cubesize);
\draw[gray,dashed] (\cubeoffset,\cubeoffset,\cubeoffset) ++(0,-\cubesize,-\cubesize-\projoffset) -- ++(0,\cubesize,0) ;

\fill[green!80] (\cubeoffset,\cubeoffset,\cubeoffset) ++(-.5*\cubesize-\projoffset,0,0) -- ++(0,0,-\rad) -- ++(0,-\rad,0) -- ++(0,0,\rad) -- cycle;
\fill[green!80] (\cubeoffset,\cubeoffset,\cubeoffset) ++(-.5*\cubesize-\projoffset,0,-\cubesize+\rad) -- ++(0,0,-\rad) -- ++(0,-\rad,0) -- ++(0,0,\rad) -- cycle;
\fill[green!80] (\cubeoffset,\cubeoffset,\cubeoffset) ++(-.5*\cubesize+\rad,-\cubesize-\projoffset,0) -- ++(0,0,-\rad) -- ++(-\rad,0,0) -- ++(0,0,\rad) -- cycle;
\fill[green!80] (\cubeoffset,\cubeoffset,\cubeoffset) ++(-.5*\cubesize-\projoffset,-\cubesize+\rad,0) -- ++(0,0,-\rad) -- ++(0,-\rad,0) -- ++(0,0,\rad) -- cycle;
\fill[green!80] (\cubeoffset,\cubeoffset,\cubeoffset) ++(-.5*\cubesize+\rad,0,-\cubesize-\projoffset) -- ++(-\rad,0,0) -- ++(0,-\rad,0) -- ++(\rad,0,0) -- cycle;
\draw[latex-,ultra thick,red] (0.05+\cubeoffset-.5*\cubesize-\projoffset,\cubeoffset-.5*\rad,\cubeoffset+.5*\rad-\cubesize) -- ++(0.6,0.0,0.0);
\draw[latex-,ultra thick,red] (0.05+\cubeoffset-.5*\cubesize-\projoffset,\cubeoffset-\cubesize+.5*\rad,\cubeoffset-.5*\rad) -- ++(0.6,0.0,0.0);
\node[text width=4cm,text centered] at (\cubeoffset+0.2,-\cubeoffset-\projoffset-0.5,\cubeoffset) (a) {\footnotesize \begin{verse}\baselineskip=10pt cropped MIP \\ annotations\end{verse}};

\pgfmathsetmacro{\xcubeoffset}{3.1}
\filldraw[fill=blue!50,draw=blue] (\xcubeoffset-.5*\cubesize+\rad,\cubeoffset,\cubeoffset) -- ++(-\rad,0,0) -- ++(0,-\rad,0) -- ++(\rad,0,0) -- cycle;
\filldraw[fill=blue!50,draw=blue] (\xcubeoffset-.5*\cubesize+\rad,\cubeoffset,\cubeoffset) -- ++(-\rad,0,0) -- ++(0,0,-\rad) -- ++(\rad,0,0) -- cycle;
\filldraw[fill=blue!50,draw=blue] (\xcubeoffset-.5*\cubesize+\rad,\cubeoffset,\cubeoffset) -- ++(0,0,-\rad) -- ++(0,-\rad,0) -- ++(0,0,\rad) -- cycle;
%\draw[gray!80,dashed] (\cubeoffset-\rad,\cubeoffset-\rad,\cubeoffset-\rad) circle [radius=\rad];
\draw[gray] (\xcubeoffset,\cubeoffset,\cubeoffset) -- ++(-.5*\cubesize,0,0) -- ++(0,-\cubesize,0) -- ++(.5*\cubesize,0,0); % -- cycle;
\draw[gray] (\xcubeoffset,\cubeoffset,\cubeoffset) -- ++(-.5*\cubesize,0,0) -- ++(0,0,-\cubesize) -- ++(.5*\cubesize,0,0); % -- cycle;
\draw[gray,dashed] (\xcubeoffset,\cubeoffset,\cubeoffset) -- ++(0,0,-\cubesize) -- ++(0,-\cubesize,0) -- ++(0,0,\cubesize) -- cycle;
\node[text width=4cm,text centered] at (\xcubeoffset-0.1,-\cubeoffset-\projoffset-0.5,\cubeoffset) (b) {\footnotesize visual hull };
%\draw[gray,dashed] (\cubeoffset,\cubeoffset,\cubeoffset) ++(-\cubesize,0,-\cubesize) -- ++(0,-\cubesize,0) -- ++(\cubesize,0,0) ++(-\cubesize,0,0) -- ++(0,0,\cubesize);
\pgfmathsetmacro{\arrowoffsetx}{1.5*\cubeoffset+0.1}
\pgfmathsetmacro{\arrowoffsety}{0.5*\cubeoffset}
\pgfmathsetmacro{\arrowoffsetz}{0}
\pgfmathsetmacro{\projoffset}{.25}
\pgfmathsetmacro{\xcubeoffset}{5.6}
\fill[black!20] (\arrowoffsetx,\arrowoffsety,\arrowoffsetz) -- ++(.25*\cubesize,0,0) -- ++(0,.125*\cubesize,0) -- ++(.25*\cubesize,-\arrowoffsety-.125*\cubesize,0) -- ++(-.25*\cubesize,-\arrowoffsety-.125*\cubesize,0) -- ++(0,.125*\cubesize,0) -- ++(-.25*\cubesize,0,0) -- cycle;
\draw[gray] (\xcubeoffset,\cubeoffset,\cubeoffset) ++(-0.5*\cubesize-\projoffset,0,0) -- ++(0,0,-\cubesize) -- ++(0,-\cubesize,0) -- ++(0,0,\cubesize) -- cycle;
\draw[gray] (\xcubeoffset,\cubeoffset,\cubeoffset) ++(0,-\cubesize-\projoffset,0) ++(0,0,-\cubesize) -- ++(-0.5*\cubesize,0,0) -- ++(0,0,\cubesize) -- ++(0.5*\cubesize,0,0);
\draw[gray] (\xcubeoffset,\cubeoffset,\cubeoffset) ++(0,0,-\cubesize-\projoffset) -- ++(-.5*\cubesize,0,0) -- ++(0,-\cubesize,0) -- ++(.5*\cubesize,0,0) ;
\draw[gray,dashed] (\xcubeoffset,\cubeoffset,\cubeoffset) ++(0,-\cubesize-\projoffset,0) -- ++(0,0,-\cubesize);
\draw[gray,dashed] (\xcubeoffset,\cubeoffset,\cubeoffset) ++(0,-\cubesize,-\cubesize-\projoffset) -- ++(0,\cubesize,0) ;

\fill[red!40] (\xcubeoffset,\cubeoffset,\cubeoffset) ++(-.5*\cubesize-\projoffset,0,0) -- ++(0,0,-\rad) -- ++(0,-\rad,0) -- ++(0,0,\rad) -- cycle;
\fill[red!40] (\xcubeoffset,\cubeoffset,\cubeoffset) ++(-.5*\cubesize+\rad,-\cubesize-\projoffset,0) -- ++(0,0,-\rad) -- ++(-\rad,0,0) -- ++(0,0,\rad) -- cycle;
\fill[red!40] (\xcubeoffset,\cubeoffset,\cubeoffset) ++(-.5*\cubesize+\rad,0,-\cubesize-\projoffset) -- ++(-\rad,0,0) -- ++(0,-\rad,0) -- ++(\rad,0,0) -- cycle;
\draw[latex-,ultra thick,red] (0.05+\xcubeoffset-.5*\cubesize-\projoffset,\cubeoffset-.5*\rad,\cubeoffset+.5*\rad-\cubesize) -- ++(0.6,0.0,0.0);
\draw[latex-,ultra thick,red] (0.05+\xcubeoffset-.5*\cubesize-\projoffset,\cubeoffset-\cubesize+.5*\rad,\cubeoffset-.5*\rad) -- ++(0.6,0.0,0.0);
\node[text width=3.3cm,text centered] at (\xcubeoffset+0.1,-\cubeoffset-\projoffset-0.5,\cubeoffset) (c) {\footnotesize \begin{verse}\baselineskip=10pt visual hull \\ projection\end{verse}};

\pgfmathsetmacro{\arrowoffsetx}{1.5*\cubeoffset+2.35}
\pgfmathsetmacro{\arrowoffsety}{0.5*\cubeoffset}
\pgfmathsetmacro{\arrowoffsetz}{0}
\pgfmathsetmacro{\projoffset}{.25}
\pgfmathsetmacro{\xcubeoffset}{5}
\fill[black!20] (\arrowoffsetx,\arrowoffsety,\arrowoffsetz) -- ++(.25*\cubesize,0,0) -- ++(0,.125*\cubesize,0) -- ++(.25*\cubesize,-\arrowoffsety-.125*\cubesize,0) -- ++(-.25*\cubesize,-\arrowoffsety-.125*\cubesize,0) -- ++(0,.125*\cubesize,0) -- ++(-.25*\cubesize,0,0) -- cycle;
%\filldraw[fill=gray!50,draw=gray] (\xcubeoffset-.5*\cubesize+\rad,\cubeoffset,\cubeoffset) -- ++(-\rad,0,0) -- ++(0,-\rad,0) -- ++(\rad,0,0) -- cycle;
%\filldraw[fill=gray!50,draw=gray] (\xcubeoffset-.5*\cubesize+\rad,\cubeoffset,\cubeoffset) -- ++(-\rad,0,0) -- ++(0,0,-\rad) -- ++(\rad,0,0) -- cycle;
%\filldraw[fill=gray!50,draw=gray] (\xcubeoffset-.5*\cubesize+\rad,\cubeoffset,\cubeoffset) -- ++(0,0,-\rad) -- ++(0,-\rad,0) -- ++(0,0,\rad) -- cycle;
%\draw[gray!80,dashed] (\cubeoffset-\rad,\cubeoffset-\rad,\cubeoffset-\rad) circle [radius=\rad];
%\draw[gray] (\xcubeoffset,\cubeoffset,\cubeoffset) -- ++(-.5*\cubesize,0,0) -- ++(0,-\cubesize,0) -- ++(.5*\cubesize,0,0); % -- cycle;
%\draw[gray] (\xcubeoffset,\cubeoffset,\cubeoffset) -- ++(-.5*\cubesize,0,0) -- ++(0,0,-\cubesize) -- ++(.5*\cubesize,0,0); % -- cycle;
%\draw[gray,dashed] (\xcubeoffset,\cubeoffset,\cubeoffset) -- ++(0,0,-\cubesize) -- ++(0,-\cubesize,0) -- ++(0,0,\cubesize) -- cycle;
\end{tikzpicture}
}
&
%\resizebox{!}{3.5cm}{
{
\begin{tikzpicture}[scale=0.8]
\clip (-0.8,-2.5) rectangle + (7.0,3.8);
\pgfmathsetmacro{\cubesize}{1.5}
\pgfmathsetmacro{\cubeoffset}{0.75}
\pgfmathsetmacro{\rad}{0.3}
\pgfmathsetmacro{\projoffset}{.25}
\draw[gray] (\cubeoffset,\cubeoffset,\cubeoffset) ++(-.5*\cubesize-\projoffset,0,0) -- ++(0,0,-\cubesize) -- ++(0,-\cubesize,0) -- ++(0,0,\cubesize) -- cycle;
\draw[gray] (\cubeoffset,\cubeoffset,\cubeoffset) ++(-.5*\cubesize,-\cubesize-\projoffset,0) -- ++(.5*\cubesize,0,0) -- ++(0,0,-\cubesize) -- ++(-.5*\cubesize,0,0) ;
\draw[gray,dashed] (\cubeoffset,\cubeoffset,\cubeoffset) ++(-.5*\cubesize,-\cubesize-\projoffset,0) -- ++(0,0,-\cubesize) ;
\draw[gray] (\cubeoffset,\cubeoffset,\cubeoffset) ++(-.5*\cubesize,0,-\cubesize-\projoffset) -- ++(.5*\cubesize,0,0) -- ++(0,-\cubesize,0) -- ++(-.5*\cubesize,0,0) ;
\draw[gray,dashed] (\cubeoffset,\cubeoffset,\cubeoffset) ++(-.5*\cubesize,0,-\cubesize-\projoffset) -- ++(0,-\cubesize,0) ;

\fill[green!80] (\cubeoffset,\cubeoffset,\cubeoffset) ++(-\cubesize-\projoffset+.5*\cubesize,0,0) -- ++(0,0,-\rad) -- ++(0,-\rad,0) -- ++(0,0,\rad) -- cycle;
\fill[green!80] (\cubeoffset,\cubeoffset,\cubeoffset) ++(0,-\cubesize-\projoffset,0) -- ++(0,0,-\rad) -- ++(-\rad,0,0) -- ++(0,0,\rad) -- cycle;
\fill[green!80] (\cubeoffset,\cubeoffset,\cubeoffset) ++(0,0,-\cubesize-\projoffset) -- ++(-\rad,0,0) -- ++(0,-\rad,0) -- ++(\rad,0,0) -- cycle;
\fill[green!80] (\cubeoffset,\cubeoffset,\cubeoffset) ++(-\cubesize-\projoffset+.5*\cubesize,0,-\cubesize+\rad) -- ++(0,0,-\rad) -- ++(0,-\rad,0) -- ++(0,0,\rad) -- cycle;
%\fill[green!80] (\cubeoffset,\cubeoffset,\cubeoffset) ++(-\cubesize+\rad,-\cubesize-\projoffset,0) -- ++(0,0,-\rad) -- ++(-\rad,0,0) -- ++(0,0,\rad) -- cycle;
\fill[green!80] (\cubeoffset,\cubeoffset,\cubeoffset) ++(0,-\cubesize+\rad,-\cubesize-\projoffset) -- ++(-\rad,0,0) -- ++(0,-\rad,0) -- ++(\rad,0,0) -- cycle;
\fill[green!80] (\cubeoffset,\cubeoffset,\cubeoffset) ++(-\cubesize-\projoffset+.5*\cubesize,-\cubesize+\rad,0) -- ++(0,0,-\rad) -- ++(0,-\rad,0) -- ++(0,0,\rad) -- cycle;
%\fill[green!80] (\cubeoffset,\cubeoffset,\cubeoffset) ++(0,-\cubesize-\projoffset,-\cubesize+\rad) -- ++(0,0,-\rad) -- ++(-\rad,0,0) -- ++(0,0,\rad) -- cycle;
%\fill[gray!80] (\cubeoffset,\cubeoffset,\cubeoffset) ++(-\cubesize+\rad,0,-\cubesize-\projoffset) -- ++(-\rad,0,0) -- ++(0,-\rad,0) -- ++(\rad,0,0) -- cycle;
\node[text width=4cm,text centered] at (\cubeoffset+0.2,-\cubeoffset-\projoffset-0.5,\cubeoffset) (a) {\footnotesize \begin{verse}\baselineskip=10pt cropped MIP \\ annotations\end{verse}};

\pgfmathsetmacro{\xcubeoffset}{3.1}
\filldraw[fill=blue!50,draw=blue] (\xcubeoffset,\cubeoffset,\cubeoffset-\cubesize+\rad) -- ++(-\rad,0,0) -- ++(0,-\rad,0) -- ++(\rad,0,0) -- cycle;
\filldraw[fill=blue!50,draw=blue] (\xcubeoffset,\cubeoffset,\cubeoffset-\cubesize+\rad) -- ++(-\rad,0,0) -- ++(0,0,-\rad) -- ++(\rad,0,0) -- cycle;
\filldraw[fill=blue!50,draw=blue] (\xcubeoffset,\cubeoffset,\cubeoffset-\cubesize+\rad) -- ++(0,0,-\rad) -- ++(0,-\rad,0) -- ++(0,0,\rad) -- cycle;
%\filldraw[fill=blue!50,draw=blue] (\cubeoffset-\cubesize+\rad,\cubeoffset,\cubeoffset) -- ++(-\rad,0,0) -- ++(0,-\rad,0) -- ++(\rad,0,0) -- cycle;
%\filldraw[fill=blue!50,draw=blue] (\cubeoffset-\cubesize+\rad,\cubeoffset,\cubeoffset) -- ++(-\rad,0,0) -- ++(0,0,-\rad) -- ++(\rad,0,0) -- cycle;
%\filldraw[fill=blue!50,draw=blue] (\cubeoffset-\cubesize+\rad,\cubeoffset,\cubeoffset) -- ++(0,0,-\rad) -- ++(0,-\rad,0) -- ++(0,0,\rad) -- cycle;
\filldraw[fill=blue!50,draw=blue] (\xcubeoffset,\cubeoffset-\cubesize+\rad,\cubeoffset) -- ++(-\rad,0,0) -- ++(0,-\rad,0) -- ++(\rad,0,0) -- cycle;
\filldraw[fill=blue!50,draw=blue] (\xcubeoffset,\cubeoffset-\cubesize+\rad,\cubeoffset) -- ++(-\rad,0,0) -- ++(0,0,-\rad) -- ++(\rad,0,0) -- cycle;
\filldraw[fill=blue!50,draw=blue] (\xcubeoffset,\cubeoffset-\cubesize+\rad,\cubeoffset) -- ++(0,0,-\rad) -- ++(0,-\rad,0) -- ++(0,0,\rad) -- cycle;
\filldraw[fill=blue!50,draw=blue] (\xcubeoffset,\cubeoffset,\cubeoffset) -- ++(-\rad,0,0) -- ++(0,-\rad,0) -- ++(\rad,0,0) -- cycle;
\filldraw[fill=blue!50,draw=blue] (\xcubeoffset,\cubeoffset,\cubeoffset) -- ++(-\rad,0,0) -- ++(0,0,-\rad) -- ++(\rad,0,0) -- cycle;
\filldraw[fill=blue!50,draw=blue] (\xcubeoffset,\cubeoffset,\cubeoffset) -- ++(0,0,-\rad) -- ++(0,-\rad,0) -- ++(0,0,\rad) -- cycle;
%\draw[gray!80,dashed] (\cubeoffset-\rad,\cubeoffset-\rad,\cubeoffset-\rad) circle [radius=\rad];
\draw[gray] (\xcubeoffset-.5*\cubesize,\cubeoffset,\cubeoffset) -- ++(.5*\cubesize,0,0) -- ++(0,-\cubesize,0) -- ++(-.5*\cubesize,0,0) ;
\draw[gray,dashed] (\xcubeoffset-.5*\cubesize,\cubeoffset,\cubeoffset) -- ++(0,-\cubesize,0);
\draw[gray] (\xcubeoffset-.5*\cubesize,\cubeoffset,\cubeoffset) -- ++(.5*\cubesize,0,0) -- ++(0,0,-\cubesize) -- ++(-.5*\cubesize,0,0) ;
\draw[gray,dashed] (\xcubeoffset-.5*\cubesize,\cubeoffset,\cubeoffset) -- ++(0,0,-\cubesize);
\draw[gray] (\xcubeoffset,\cubeoffset,\cubeoffset) -- ++(0,0,-\cubesize) -- ++(0,-\cubesize,0) -- ++(0,0,\cubesize) -- cycle;
%\draw[gray,dashed] (\cubeoffset,\cubeoffset,\cubeoffset) ++(-\cubesize,0,-\cubesize) -- ++(0,-\cubesize,0) -- ++(\cubesize,0,0) ++(-\cubesize,0,0) -- ++(0,0,\cubesize);
\node[text width=4cm,text centered] at (\xcubeoffset-0.1,-\cubeoffset-\projoffset-0.5,\cubeoffset) (b) {\footnotesize visual hull };
\pgfmathsetmacro{\arrowoffsetx}{1.5*\cubeoffset+0.1}
\pgfmathsetmacro{\arrowoffsety}{0.5*\cubeoffset}
\pgfmathsetmacro{\arrowoffsetz}{0}
\pgfmathsetmacro{\projoffset}{.25}
\pgfmathsetmacro{\xcubeoffset}{5.6}
\fill[black!20] (\arrowoffsetx,\arrowoffsety,\arrowoffsetz) -- ++(.25*\cubesize,0,0) -- ++(0,.125*\cubesize,0) -- ++(.25*\cubesize,-\arrowoffsety-.125*\cubesize,0) -- ++(-.25*\cubesize,-\arrowoffsety-.125*\cubesize,0) -- ++(0,.125*\cubesize,0) -- ++(-.25*\cubesize,0,0) -- cycle;
\draw[gray] (\xcubeoffset,\cubeoffset,\cubeoffset) ++(-.5*\cubesize-\projoffset,0,0) -- ++(0,0,-\cubesize) -- ++(0,-\cubesize,0) -- ++(0,0,\cubesize) -- cycle;
\draw[gray] (\xcubeoffset,\cubeoffset,\cubeoffset) ++(-.5*\cubesize,-\cubesize-\projoffset,0) -- ++(.5*\cubesize,0,0) -- ++(0,0,-\cubesize) -- ++(-.5*\cubesize,0,0) ;
\draw[gray,dashed] (\xcubeoffset,\cubeoffset,\cubeoffset) ++(-.5*\cubesize,-\cubesize-\projoffset,0) -- ++(0,0,-\cubesize) ;
\draw[gray] (\xcubeoffset,\cubeoffset,\cubeoffset) ++(-.5*\cubesize,0,-\cubesize-\projoffset) -- ++(.5*\cubesize,0,0) -- ++(0,-\cubesize,0) -- ++(-.5*\cubesize,0,0) ;
\draw[gray,dashed] (\xcubeoffset,\cubeoffset,\cubeoffset) ++(-.5*\cubesize,0,-\cubesize-\projoffset) -- ++(0,-\cubesize,0) ;

\fill[red!40] (\xcubeoffset,\cubeoffset,\cubeoffset) ++(-\cubesize-\projoffset+.5*\cubesize,0,0) -- ++(0,0,-\rad) -- ++(0,-\rad,0) -- ++(0,0,\rad) -- cycle;
\fill[red!40] (\xcubeoffset,\cubeoffset,\cubeoffset) ++(0,-\cubesize-\projoffset,0) -- ++(0,0,-\rad) -- ++(-\rad,0,0) -- ++(0,0,\rad) -- cycle;
\fill[red!40] (\xcubeoffset,\cubeoffset,\cubeoffset) ++(0,0,-\cubesize-\projoffset) -- ++(-\rad,0,0) -- ++(0,-\rad,0) -- ++(\rad,0,0) -- cycle;
\fill[red!40] (\xcubeoffset,\cubeoffset,\cubeoffset) ++(-\cubesize-\projoffset+.5*\cubesize,0,-\cubesize+\rad) -- ++(0,0,-\rad) -- ++(0,-\rad,0) -- ++(0,0,\rad) -- cycle;
%\fill[red!40] (\xcubeoffset,\cubeoffset,\cubeoffset) ++(-\cubesize+\rad,-\cubesize-\projoffset,0) -- ++(0,0,-\rad) -- ++(-\rad,0,0) -- ++(0,0,\rad) -- cycle;
\fill[red!40] (\xcubeoffset,\cubeoffset,\cubeoffset) ++(0,-\cubesize+\rad,-\cubesize-\projoffset) -- ++(-\rad,0,0) -- ++(0,-\rad,0) -- ++(\rad,0,0) -- cycle;
\fill[red!40] (\xcubeoffset,\cubeoffset,\cubeoffset) ++(-\cubesize-\projoffset+.5*\cubesize,-\cubesize+\rad,0) -- ++(0,0,-\rad) -- ++(0,-\rad,0) -- ++(0,0,\rad) -- cycle;
\fill[red!40] (\xcubeoffset,\cubeoffset,\cubeoffset) ++(0,-\cubesize-\projoffset,-\cubesize+\rad) -- ++(0,0,-\rad) -- ++(-\rad,0,0) -- ++(0,0,\rad) -- cycle;
\node[text width=3.3cm,text centered] at (\xcubeoffset+0.1,-\cubeoffset-\projoffset-0.5,\cubeoffset) (c) {\footnotesize \begin{verse}\baselineskip=10pt visual hull \\ projection\end{verse}};
%\fill[red!40] (\xcubeoffset,\cubeoffset,\cubeoffset) ++(-\cubesize+\rad,0,-\cubesize-\projoffset) -- ++(-\rad,0,0) -- ++(0,-\rad,0) -- ++(\rad,0,0) -- cycle;
\pgfmathsetmacro{\arrowoffsetx}{1.5*\cubeoffset+2.35}
\pgfmathsetmacro{\arrowoffsety}{0.5*\cubeoffset}
\pgfmathsetmacro{\arrowoffsetz}{0}
\pgfmathsetmacro{\projoffset}{.25}
\pgfmathsetmacro{\xcubeoffset}{5}
\fill[black!20] (\arrowoffsetx,\arrowoffsety,\arrowoffsetz) -- ++(.25*\cubesize,0,0) -- ++(0,.125*\cubesize,0) -- ++(.25*\cubesize,-\arrowoffsety-.125*\cubesize,0) -- ++(-.25*\cubesize,-\arrowoffsety-.125*\cubesize,0) -- ++(0,.125*\cubesize,0) -- ++(-.25*\cubesize,0,0) -- cycle;
\end{tikzpicture}  
}
\\[-2mm]
(b)&(c)
\end{tabular}

%% file: experiments.tex
% !TEX root = top.tex
% !TEX spellcheck = en-US

\input{neuron}
\section{Experimental Evaluation \label{sec:exp}}

\newcommand{\BASE}[0]{{\bf BASE}}
\newcommand{\OURS}[0]{{\bf OURS}}
\newcommand{\OURT}[0]{{\bf OURS-2Projections}}
\newcommand{\OURU}[0]{{\bf OURS-1Projections}}
\newcommand{\NHUL}[0]{{\bf OURS-NoHull}}
\newcommand{\SMAX}[0]{{\bf OURS-SoftMax}}

\newcommand{\Axons}[0]{{Axons}}
\newcommand{\Retina}[0]{{Retina}}
\newcommand{\Angio}[0]{{Angiography}}
\newcommand{\Mouse}[0]{{Brain}}

\subsection{Datasets}

We tested our approach on four data sets that differ in terms of the imaged tissue, the acquisition modality and the image resolution. There are substantial variations between these datasets with respect to the density of the structures of interest, their appearance, and the amount of clutter originating from extraneous objects. Together, they constitute an exhaustive benchmark for 3D delineation.

\paragraph{\bf \Axons} The dataset comprises 16 stacks of 2-photon microscopy images of mouse neural tissue, with sizes ranging from $40 \times 200 \times 200$ to $136 \times 322 \times 500$ voxels and a resolution of $0.8 \times 0.26 \times 0.26$ \si{\micro\meter}. The images were acquired in vivo, from a mouse with a translucent window implanted in the scalp. We split the data into a test set of two volumes of size $136 \times 233 \times 500$, and a training set of 14 smaller volumes. The top row of Fig.~\ref{fig:neuron} depicts one of the test volumes.

\paragraph{\bf \Retina} The dataset is made of two confocal microscopy image stacks depicting retinal blood vessels. The stacks have a size of $1024 \times 1024 \times 110$ voxels and a resolution of $0.62$ \si{\micro\meter}. We use one of them for training and the other, depicted in Fig.~\ref{fig:neuron}, for testing. Since most vessels are located within a 50-pixel high XY slice, MIPs in the  X and Y directions are very cluttered. Therefore, we split the volume into 16 subvolumes, sized $256 \times 256 \times 110$ voxels, and annotated their MIPs. In other words, we also traced the vertical faces of the smaller volumes. This only requires annotating 6 additional $1024 \times 110$ images, which is still fast. The middle row of Fig.~\ref{fig:neuron} describes both our 2D annotations and the segmentation results for the test volume.

\paragraph{\bf \Angio} This set of MRI brain scans~\citep{Bullitt05}, one of which is shown in Fig.~\ref{fig:neuron}, is publicly available. It consists of 42 annotated stacks, which we cropped to a size of $416 \times 320 \times 128$ voxels by removing their empty margins. Their resolution is $0.5 \times 0.5 \times 0.6$ \si{\milli\meter}. We randomly partitioned the data into 31 training and 11 test volumes. As in the case of the retinal vessels,  we decreased the visual clutter by splitting each volume into four $208 \times 160 \times 128$ subvolumes for which we produced 2D annotations. This requires annotating an additional $416 \times 128$ image and a $320 \times 128$ one for each training volume. The bottom row of Fig.~\ref{fig:neuron} describes both our 2D annotations and our results on one of the test stacks.

\paragraph{\bf \Mouse} The dataset is a part of a 2-photon microscopy scan of a whole mouse brain. It contains 14 stacks of size $250 \times 250 \times 200$ voxels and a spatial resolution of $1.0 \times 0.3 \times 0.3$ \si{\micro\meter}. Compared to the \Axons{} dataset the volumes are more diverse since they were pooled randomly from different brain regions. We use 10 stacks for training and 4 for testing. The last row of Fig.~\ref{fig:neuron} depicts an example volume.

All the manual annotations are expressed in terms of 2D and 3D centerlines of the underlying structures. We then use a pixel-width of 11 for \Axons, \Retina{} and \Mouse{} datasets, and 7 for the \Angio{} volumes, to define the area to ignore around the centerline when computing the loss, as discussed in Section~\ref{sec:2DAnnotations}, as well as to compute the visual hulls, as described in Section~\ref{sec:spaceCarving}. 

\subsection{User Study}
\label{sec:study}
%\color{blue}
The usefulness of our approach is predicated on the claim that annotating linear structures in 2D is much easier than doing it in 3D, while the two annotation types give equally good results when used for training. To substantiate this claim, we conducted a user study involving 15 PhD students used to performing such delineation for research purposes. We asked them to annotate one volume from the \Mouse{} dataset in 2D, and another one in 3D. The annotation was performed using the Fiji Simple Neurite Tracer plugin~\citep{Frangi98}. 
We present the analysis of the data collected in the study below. In subsection~\ref{sssec:efficiency} we demonstrate that switching to annotating in 2D enables annotating the data set twice as fast as in 3D. In subsection~\ref{sssec:quality} we show that, the 2D annotations are nevertheless consistent with the 3D ones. Finally, in subsection~\ref{sssec:performance} we demonstrate that, when used for training with our method, they yield networks performing on par with ones trained on the full 3D annotations.

\input{tab_times}

\subsubsection{Efficiency of MIP annotation}
\label{sssec:efficiency}
To estimate the annotation workload we recorded the wall-clock time it took the participants to complete their tasks. We present the results in Table~\ref{tab:2d3dtimes}. 
Annotating all 15 volumes in 3D took the participants of our user study 10 working hours in total. The time needed to label the dataset in 2D was 6.5 hours, or 65\% of the 3D annotation time, when annotating 3 MIPs per volume. This could further be reduced to 45\% by annotating only two projections per volume, and to 25\% by annotating only one. The differences in the average time needed to annotate each of the views stem from the non-isotropy of the data. The scans have lower resolution along the Z axis than in the XY-plane. Additionally, the sizes of the annotated volumes along these dimensions differ. We give additional details in Fig.~\ref{fig:user}. 
\input{user}

\subsubsection{Quality of the 2D annotations} %Consistency of 2D and 3D annotations}
\label{sssec:quality}
The results of our user study suggest that annotating a dataset in 2D requires two times less work than doing it in 3D. But are the 2D annotations equally good as the 3D ones:
2D projections carry less information than the original 3D data and one might wonder if this affects the quality of the 2D annotations. 
To answer this, we evaluated the quality of the 2D projection annotations produced in our user study by comparing them to the 3D annotations.
More precisely, we projected the 3D annotations and compared the 2D MIP annotations to the resulting projections.
We computed the precision $P$ and recall $R$ of the 2D annotations with respect to the projections of the 3D annotations, defined as $P_{\twoD\threeD}=\frac{\sum_{ij} [\gt^\twoD_{\Xindex\Yindex}=1][\gt^\threeD_{\Xindex\Yindex}=1]}{\sum_{ij} [\gt^\twoD_{\Xindex\Yindex}=1]}$ and $R_{\twoD\threeD}=\frac{\sum_{ij} [\gt^\twoD_{\Xindex\Yindex}=1][\gt^\threeD_{\Xindex\Yindex}=1]}{\sum_{ij} [\gt^\threeD_{\Xindex\Yindex}=1]}$, where $[\cdot]$ is the Iverson bracket, and the summation is over all pixels of the projection. We found $P=75\%$ and $R=70\%$ indicating reasonable consistency. Given that the annotations are one-pixel-thick centerlines, some of the inconsistent annotations might simply be shifted by a small distance, while others may be missing altogether. To investigate this, we checked what percentage of annotations of one type is within a distance of no more than $d$ pixels to the closest annotation of the other type. The results are presented in Fig.~\ref{fig:consistency_2D3D}. We vary $d$ between 1 and 10 and observe that over 95\% of all 2D annotations are within a distance of 3 pixels from a projection of a 3D annotation, and vice versa. The results suggest that less than 5\% of annotations of each type are inconsistent with the annotations of the other type.

\input{consistency_plots}

%\paragraph{\bf Quality of the 2D annotations - consistency across projections}
We have shown that the 2D projection annotations are roughly consistent with the 3D annotations. However, since the 2D annotations are performed independently for different projections, inconsistencies may still occur between the 2D annotations of different projections of the same volume. More precisely, each pair of projections of a 3D volume has one dimension in common. Annotations of the two projections are consistent, if for a foreground voxel of the volume, the corresponding foreground pixels in both projection annotations have the same coordinate along the common dimension. The concept is illustrated in Fig.~\ref{fig:consistency}. 
\input{consistency}
To quantify the inconsistency of the annotations resulting from our user study, we build up on the fact that a pair of isolated, inconsistent 2D annotations, like the ones presented in Fig.~\ref{fig:consistency}, creates an empty visual hull. Therefore, the number of inconsistent 2D annotations can be estimated by constructing a visual hull from the 2D projection annotations, projecting the hull back to 2D and counting the number of positive labels that fall outside of the hull projection. As in the case of the hull-based filtration introduced in section~\ref{sec:spaceCarving}, projections of the hull may fail to eliminate some inconsistent annotations, which means that the resulting estimate is a lower bound. Additionally, we can estimate the degree of inconsistency by verifying how much the position of the inconsistent annotations differs along the common dimension. That is, we estimate how many annotations are inconsistent by no more than a given distance $d$ by dilating the annotations with a structuring element of radius $d$ before constructing the hull. The results are presented in Fig.~\ref{fig:consistency_mip}. This procedure confirms that at least 20\% of the annotations are inconsistent, but almost never by more than 3 pixels.
The effect of training on inconsistent annotations is that the error signal that is focused on a single voxel when the annotations are consistent, gets distributed over a larger number of voxels.
However, as demonstrated below, the performance attained by training on MIP annotations in the experiments appears not to be affected by this level of inconsistency.

\subsubsection{Performance}
\label{sssec:performance}
We have asserted the high quality of 2D annotations, and we now confirm their utility for training a Deep Net. As stated above, even though they are highly consistent with the 3D annotations, they \emph{do} contain less information. Moreover, as explained in section~\ref{sec:2DAnnotations}, the proposed 2D loss function~\eqref{eq:loss2D} yields very sparse gradients with respect to the 3D output of the network, with only a single nonzero value in each row, column, or tube. It is not clear \emph{a priori} that such sparse error signals are equivalent to the dense gradients obtained on full 3D annotations.
To verify this, we compared performance of networks trained on the two types of annotations.
We express the performance in terms of the maximum F1 score---the harmonic mean between the precision and recall---a standard metric for binary segmentation evaluation. In Table~\ref{tab:2d3dtimes}, we report these scores when using either 3D annotations or 2D annotations, in three, two, or only one MIP. When using three, or even only two MIPs, there is virtually no performance loss for a reduction in annotation time of 36, and 55\%, respectively.
This surprisingly high performance in spite of the sparsity of gradients our method yield can be explained by analogy to space carving as mentioned in section \ref{sec:2DAnnotations}. 
The method finds its limits when we annotate only one MIP, which results in a severe performance drop. This makes intuitive sense because, for reasonably simple shapes, space carving can yield informative estimates  from only two views but not from a single one. 

%To eliminate the possibility that one kind of annotation was of higher quality than the other, in the remainder of this section, we compare them and demonstrate that they are essentially equivalent. 

%We report the results in Table~\ref{. The participants of the study required in total 10 hours of work to complete the 3D annotatins. Annotating 3 2D Maximum Intensity Projections for each volume took around 6 hours and 30 minutes, which amounts to a 35\% reduction of annotation time. Annotating only 2 MIPs per volume enables further time saving, totaling 55\% of the 3D annotation time. Finally, limiting the annotation effort to a single projection per volume could potentially lead to decreasing the annotation effort by 75\%. However, this level of economy is accompanied by a significant performance drop, as presented in section \ref{sec:3Dvs2D}.

\color{black}
\subsection{Further confirmation}
\label{sec:3Dvs2D}
\input{tab2d3d.tex}

In the user study of Section~\ref{sec:study} the 2D and 3D annotations were generated independently. We demonstrated that they were roughly equivalent, yielding networks of similar performance when used for training. 
To evaluate the proposed approach more extensively, for various imaging modalities and specimens, we perform experiments on the three remaining data sets. Instead of performing the 2D annotations from scratch, we now use projections of the 3D annotations as the annotations of 2D projections. This is not what we would do in practice but it guarantees that their quality is {\it exactly} the same, while still enabling to test the basic concept of training on less informative 2D annotations, with a loss function yielding extremely sparse gradients.
We report our results in 
Table~\ref{tab:2d3d}.
%In Table~\ref{tab:2d3d}, we compare results obtained by training either on 2D or on 3D annotations in terms of the F1 score---the harmonic mean of precision and recall, which is a standard measure of binary segmentation performance---computed in 3D with respect to the 3D annotations. To ensure that the scores are comparable in both scenarios, we use the projections of the 3D annotations as our 2D annotations. 
In the rightmost column, we give an estimate of the time saved by  generating the 2D annotations instead of the 3D ones on the basis of the above user study. When training on 3 or 2 MIPs per volume, we obtain roughly the same results as when training on full 3D annotations---slightly better for the \Axons{} and \Mouse{}, and slightly worse for the \Retina{} and \Angio{} datasets---while, as shown in section \ref{sec:study}, the corresponding annotation effort is decreased by 45 and 55 percent, respectively. Summarizing, training on 2 MIP annotations per volume enables attaining the same precision as training on the full 3D annotations, but at half of the annotation cost. These results are fully consistent with the findings of the user study presented above. While offering further time saving, the reduction of the amount of annotations used to a single projection per volume leads to a substantial performance drawback. We leave it for future work to investigate possible methods of preventing this adverse effect.

%In experiments performed so far, projections of 3D annotations were used as MIP annotations. This enables a fair comparison of performance of the networks trained on MIP annotations and on the full 3D annotations, as test performance is evaluated in terms of consistency with the full 3D annotations for both of these setups. However, it is not clear {\it a priori} if annotations performed from scratch on 2D MIPs are as correct and complete as the projections of 3D annotations. It is also not clear, how possible differences affect performance of the trained network. To verify that, we use the 2D MIP annotations produced during the user study for the \Mouse{} data set to train a deep network. The results are presented in Table~\ref{tab:2d3d}. The performance of the network trained on the MIP annotations matches the performance attained in result of training on 2D projections of the full 3D annotations.

Whether using 3D or 2D annotations, these results rely on the modified U-Net architecture discussed in Section~\ref{sec:impl}. For completeness, we also list in Table~\ref{tab:2d3d} the performance of three earlier methods. One alternative method of limiting annotation effort required to train a volumetric Deep Net is to annotate a small subset of slices of the original volume~\citep{Cicek16}. In our experiments, the number of annotated slices used to train the network using this approach exactly matched the number of projections used in our method. For a fair comparison, we also used the same network architecture in the two sets of experiments. While for the \Retina{} and \Angio{} datasets the performance of a network trained on slice annotations closely matched that of the network trained on MIP annotations, the performance gap is larger for the two datasets depicting neural tissue. Moreover, it is often the case that the topology of linear structures is more easily disambiguated in the projections than in isolated slices, which makes annotating the projections easier. We also compare the performance of our method to a hand-crafted tubular structures detector \citep{Turetken13c} that does not require any annotations. Not surprisingly, it performs well on the \Retina{} dataset, used by the authors to develop the technique, but fails to generalize to the other datasets, not considered when designing the detector. The last baseline used in the experiments is a regression-based approach to delineation \citep{Sironi16a}, trained on the original set of 3D annotations, which our approach also outperforms.

%% file: neuron.tex
% !TEX root = ../main.tex
% !TEX spellcheck = en-US

\begin{figure*}
\centering
\setlength{\tabcolsep}{4pt}
\footnotesize
\begin{tabular}{m{.02\textwidth}>{\centering\arraybackslash}m{.3\textwidth}>{\centering\arraybackslash}m{.3\textwidth}>{\centering\arraybackslash}m{.3\textwidth}}
I &
\includegraphics[angle=90, width=0.25\textwidth]{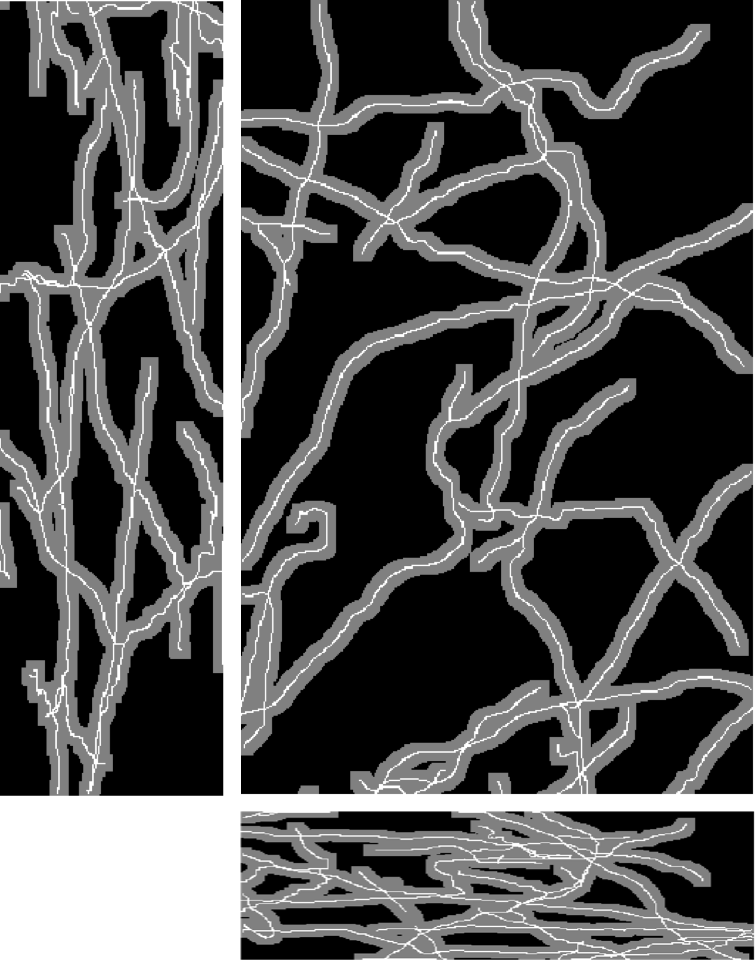}&
\includegraphics[width=0.25\textwidth]{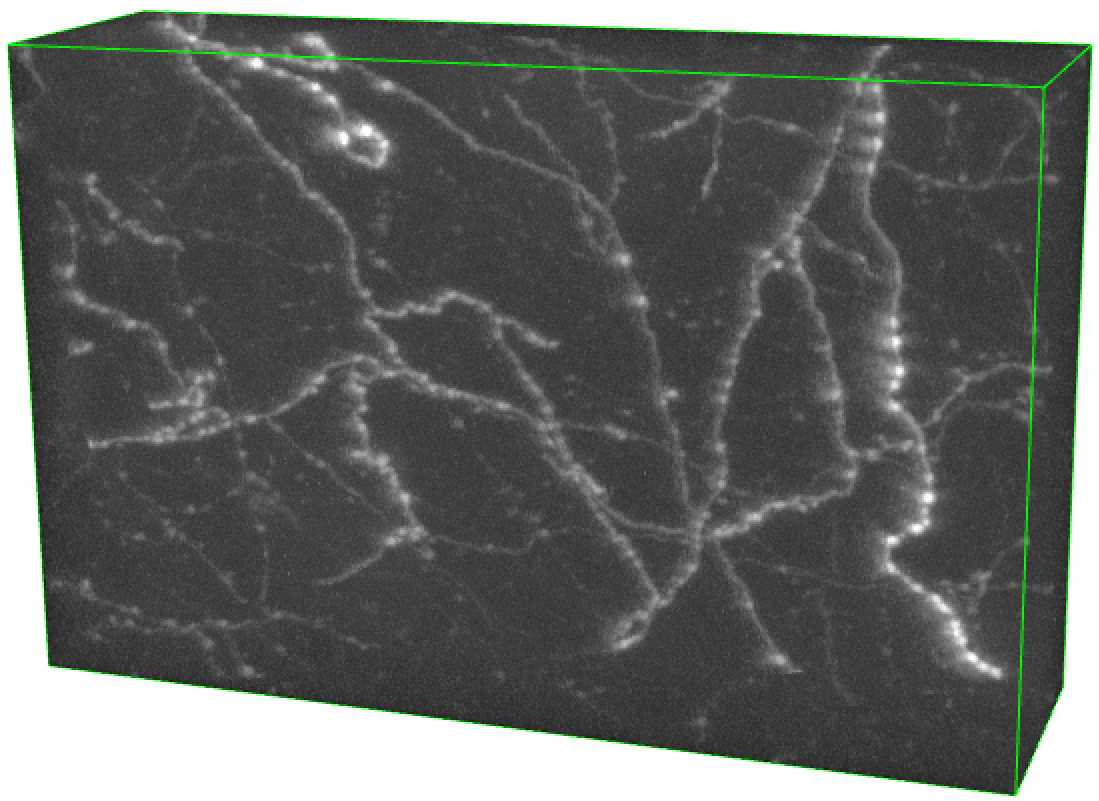}&
\includegraphics[width=0.25\textwidth]{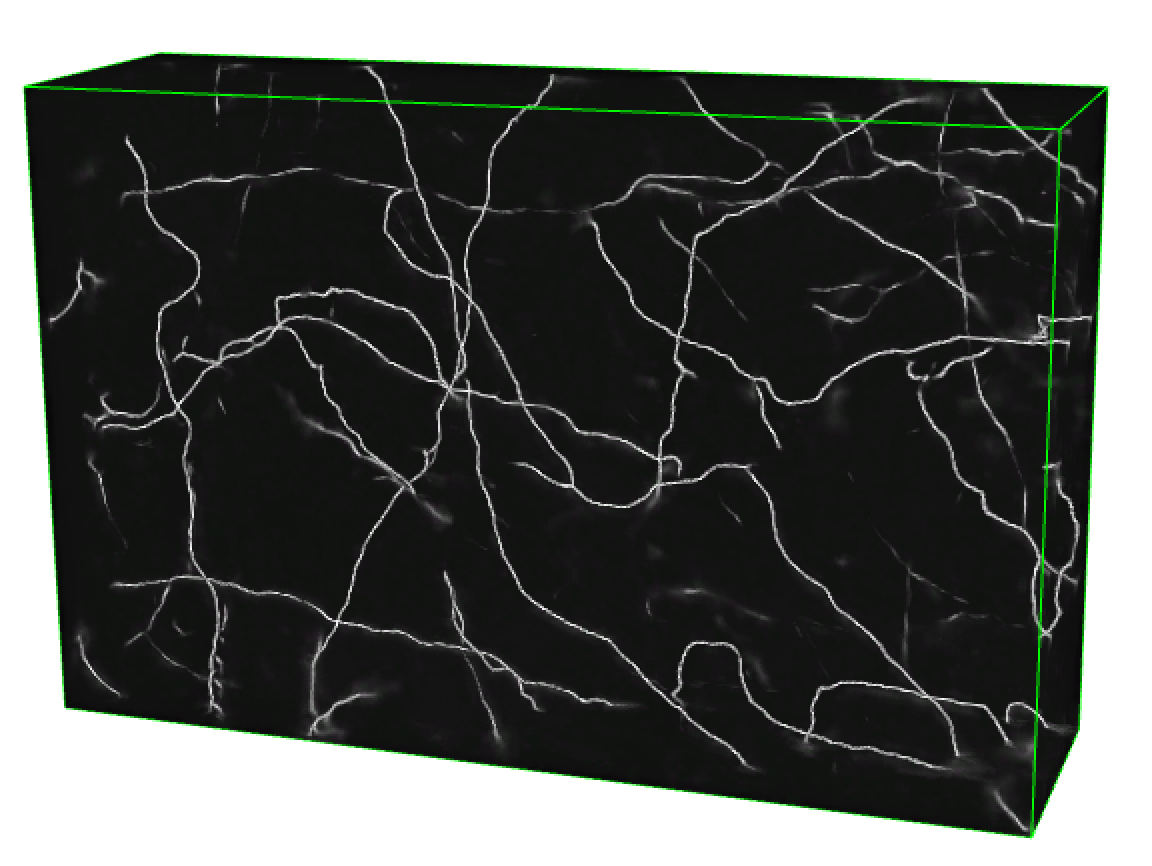}\\
II &
\includegraphics[angle=90, width=0.25\textwidth]{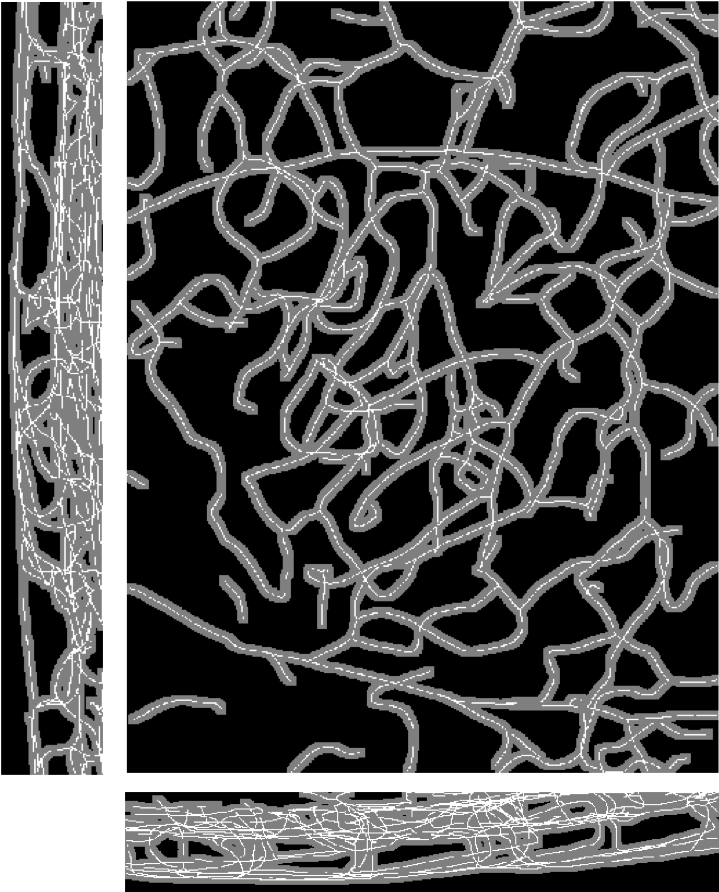}&
\includegraphics[width=0.25\textwidth]{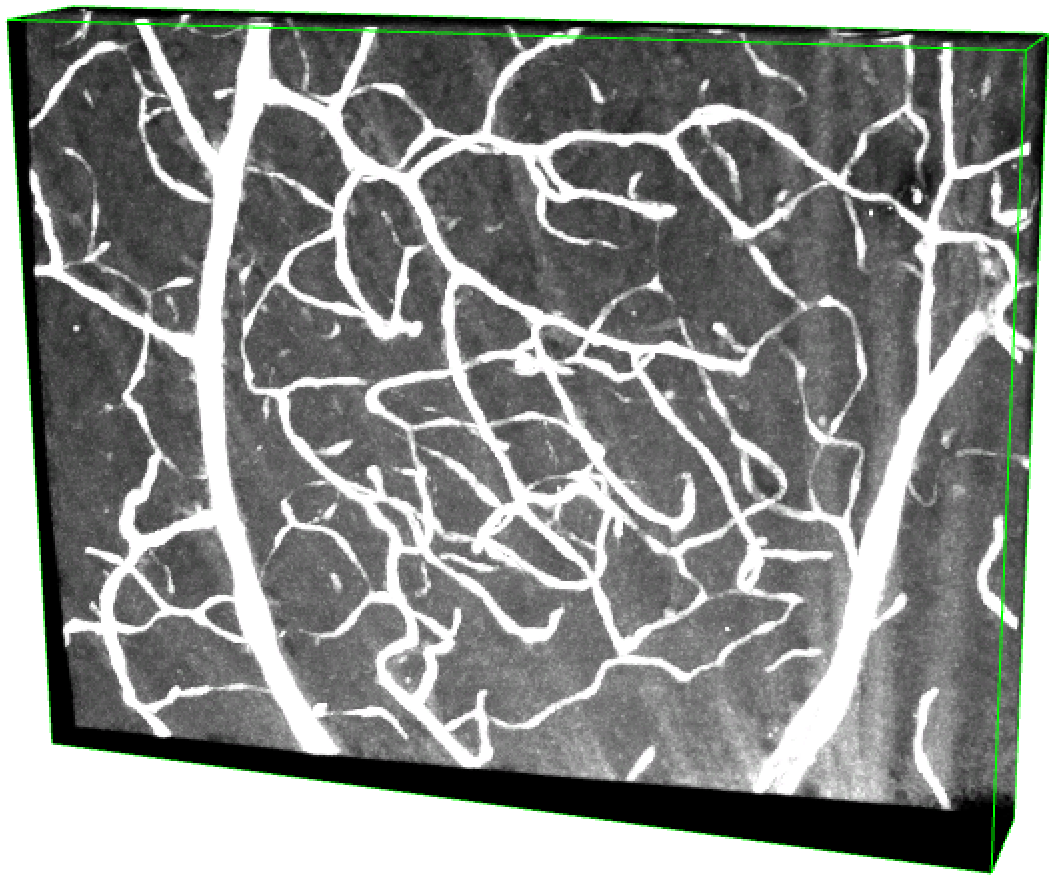}&
\includegraphics[width=0.25\textwidth]{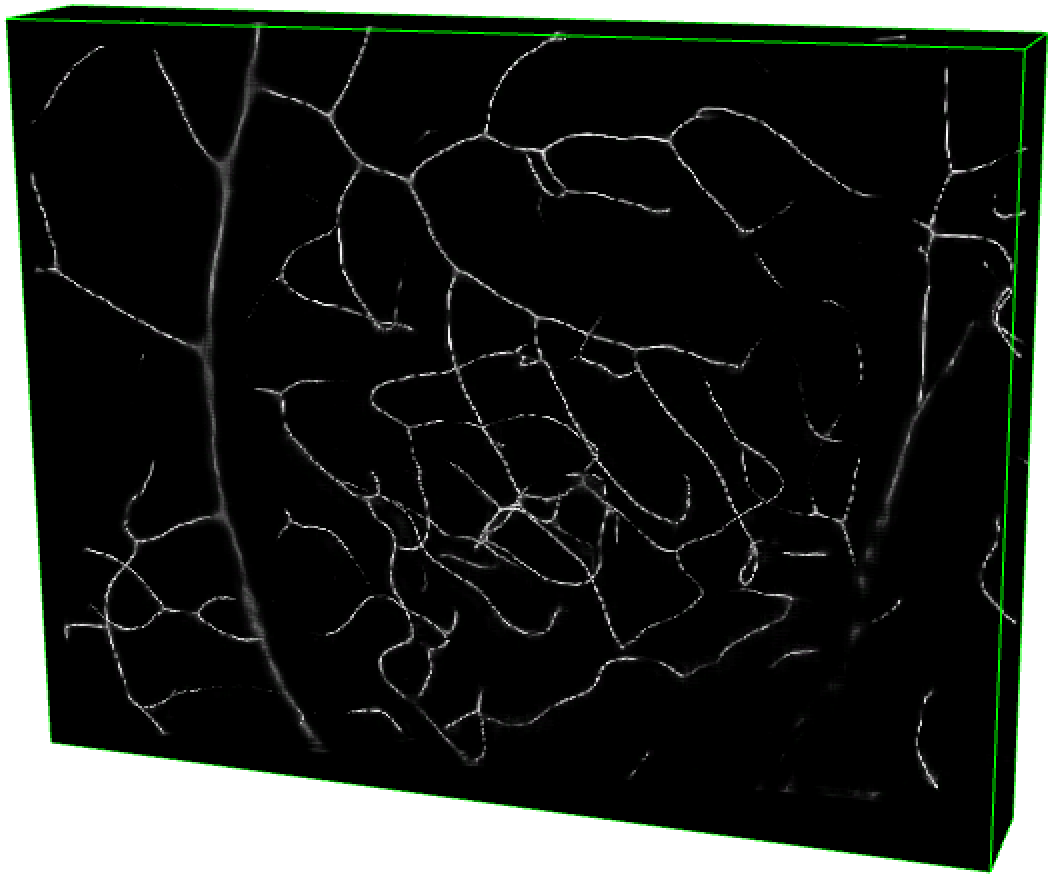}\\
III &
\includegraphics[angle=90, width=0.25\textwidth]{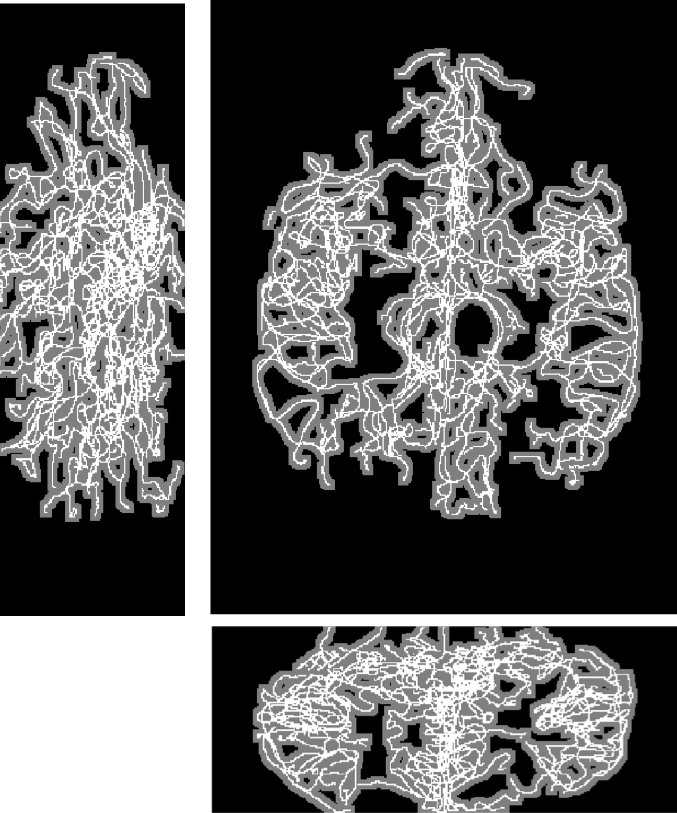}&
\includegraphics[width=0.25\textwidth]{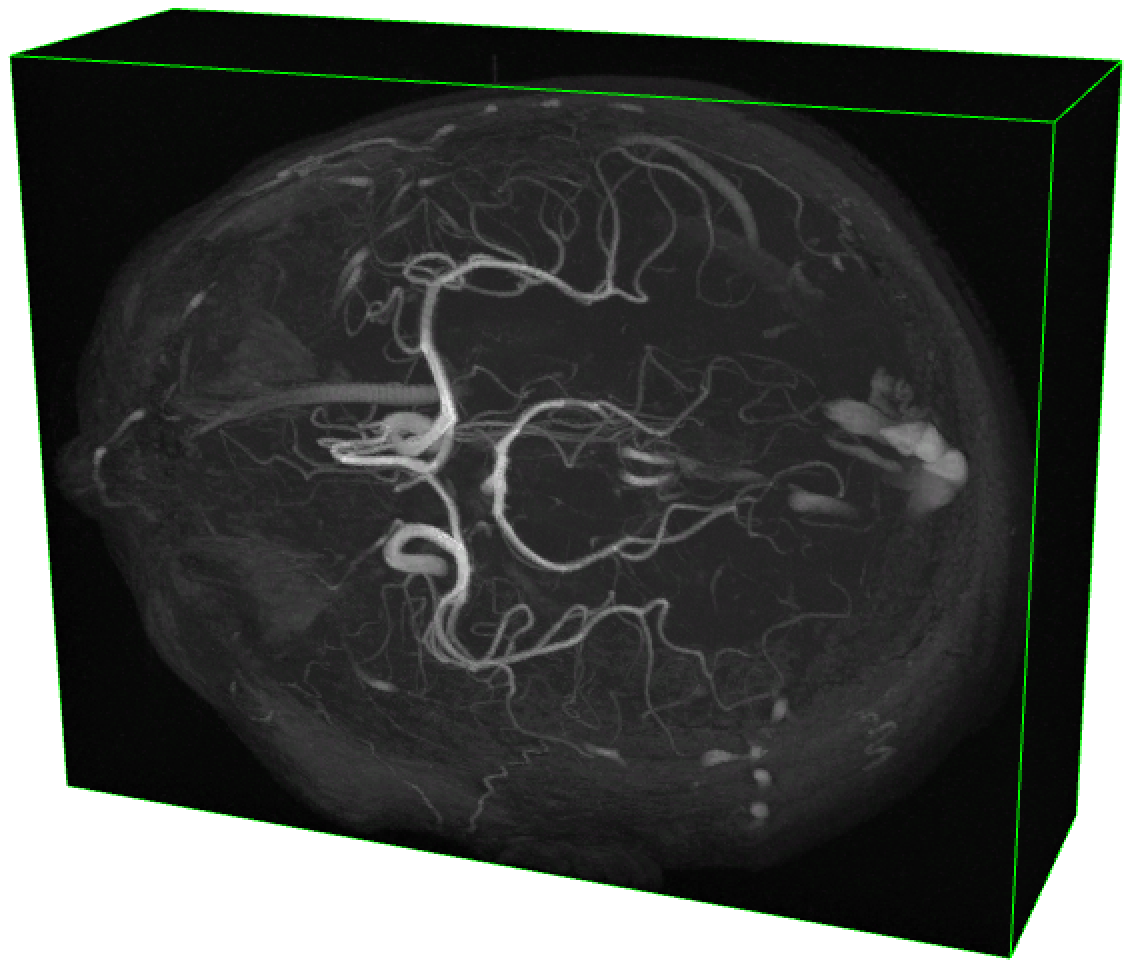}&
\includegraphics[width=0.25\textwidth]{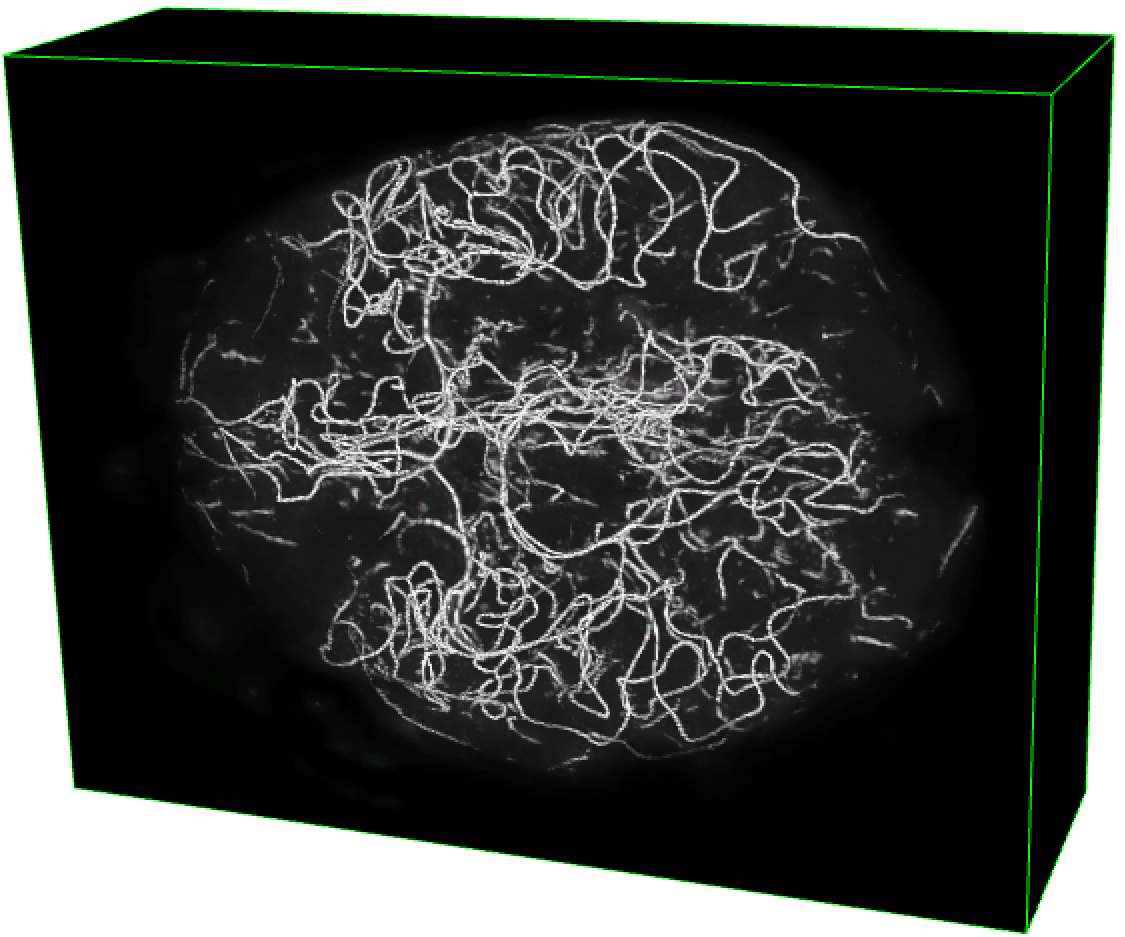}\\
IV  &
\includegraphics[width=0.25\textwidth]{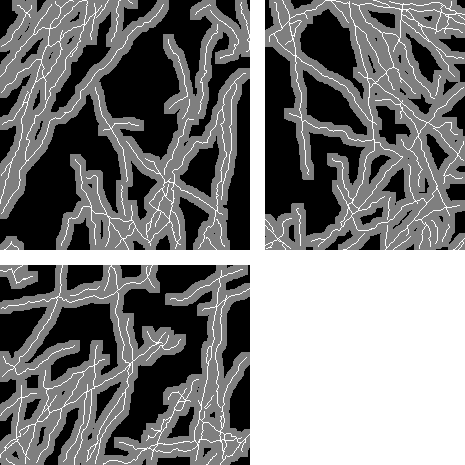}&
\includegraphics[width=0.25\textwidth]{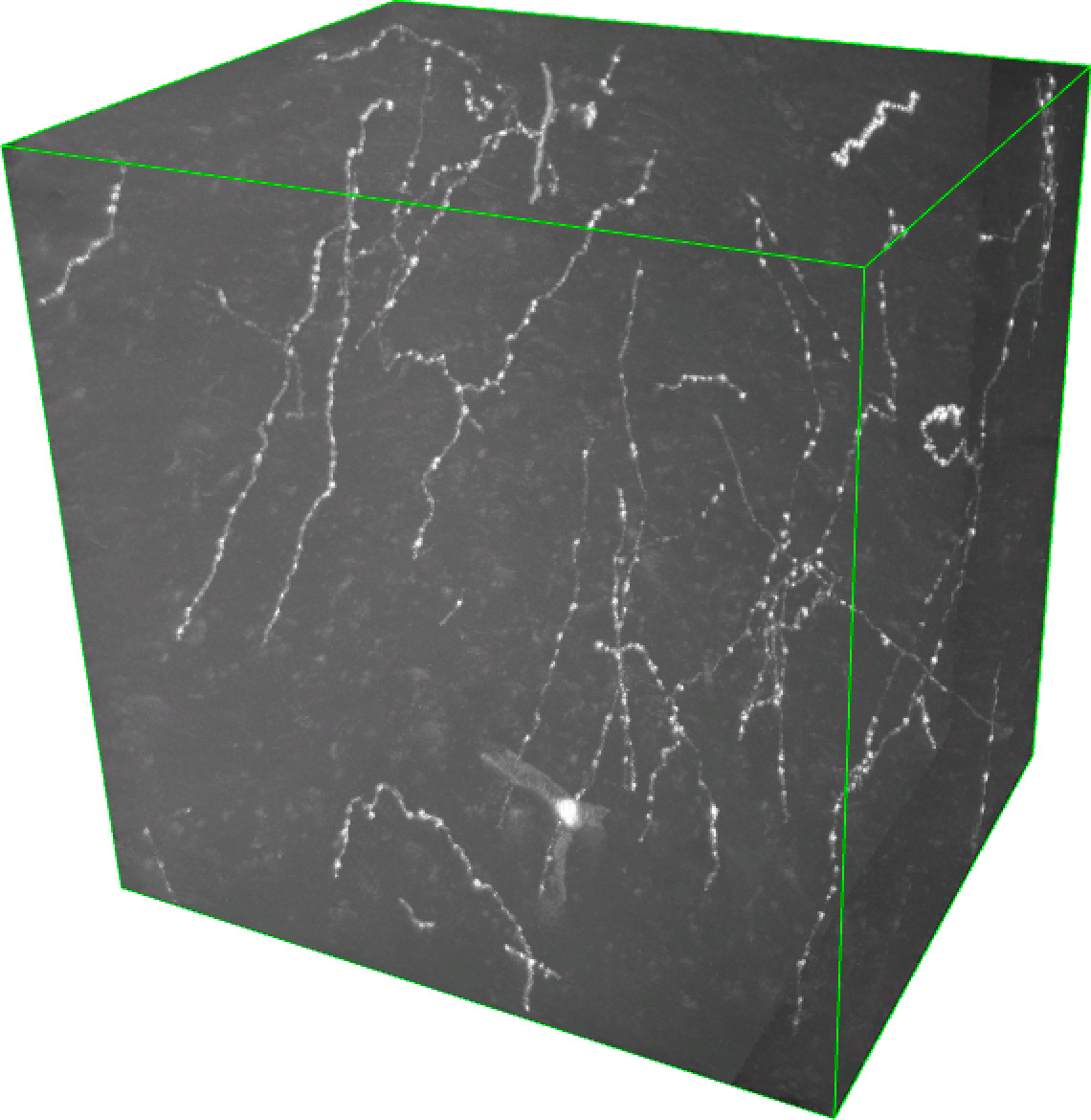}&
\includegraphics[width=0.25\textwidth]{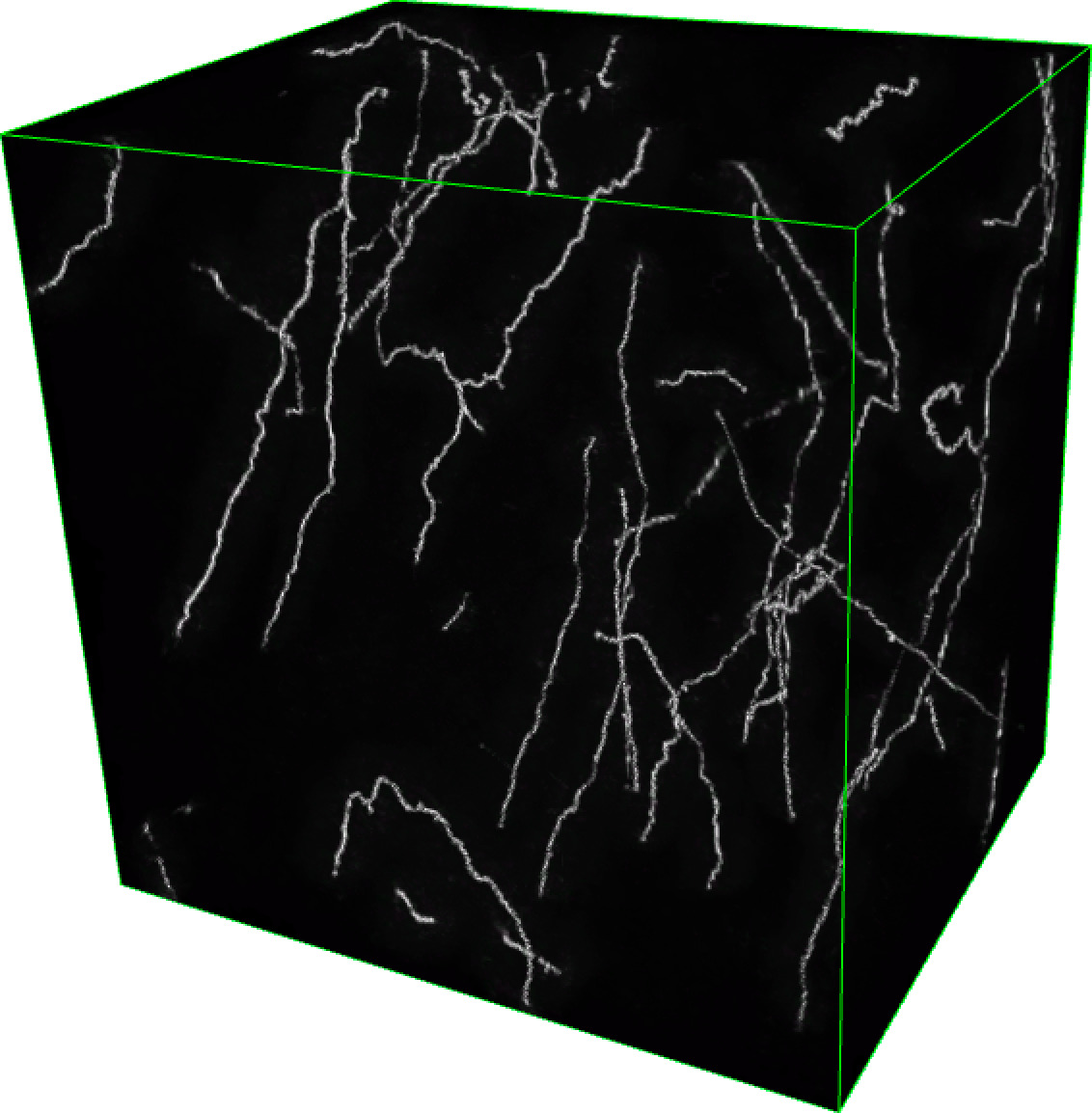}\\
 &(a)&(b)&(c)
\end{tabular}
\vspace{-3mm}
\caption{Results on our four datasets, from top to bottom, axons, retinal blood vessels, brain vasculature in MRA scans, and neural tissue in mouse brain. (a)  2D annotations in 3 MIPs of a test volume. The foreground centerline annotations are marked in white and the regions to be ignored around them in gray.  (b) Input test image volume. (c) Output segmentation. }
\label{fig:neuron}
\end{figure*}

%% file: tab_times.tex
% !TEX root = ../main.tex
% !TEX spellcheck = en-US

\begin{table}
\footnotesize
\newcommand{\fxw}[1]{\makebox[\widthof{16.0}][r]{#1}}
\setlength{\tabcolsep}{5pt}
\centering
\caption{The total time needed to complete annotations of the whole Mouse Brain dataset in the user study. \label{tab:2d3dtimes}}
\begin{tabular}{lccc}
% & \multicolumn{6}{c}{F1 score [\%] and  diff.\ to NN/3D annot.}& est.\ time gain [\%] \\
% & \multicolumn{1}{c}{axons} & \multicolumn{2}{c}{retina} &  \multicolumn{2}{c}{brain} & \\
Annotation method  & \multicolumn{2}{c}{Annotation time} & Performance\\
& [min]& [\% 3D time] & [F1 score]\\
\hline
 Annotating in 3D     &  609 & \fxw{100} & 80.2 \\
 Annotating 3 2D MIPs &  387 & \fxw{64}  & 80.0 \\
 Annotating 2 2D MIPs &  277 & \fxw{45}  & 80.0 \\
 Annotating 1 2D MIP  &  152 & \fxw{25}  & 49.2 \\
\hline \\[-2mm]
\end{tabular}
\end{table}

%% file: user.tex
% !TEX root = ../main.tex
% !TEX spellcheck = en-US

\pgfplotsset{compat = 1.3}
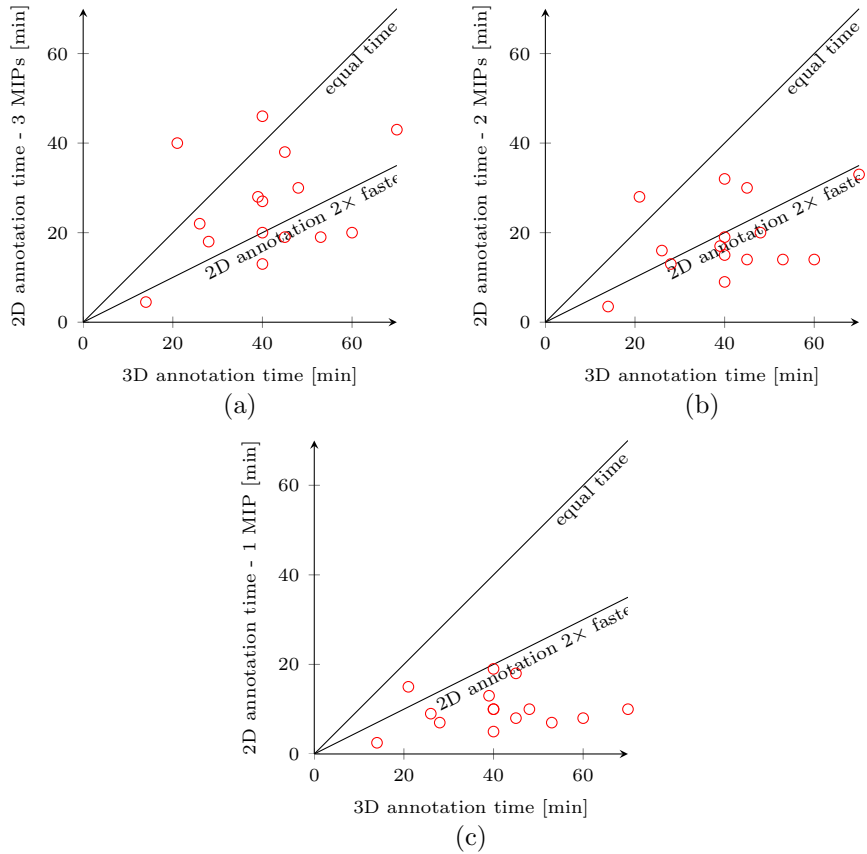
\begin{figure*}
%\setlength{\tabcolsep}{0cm}
%\begin{tabular}{>{\centering\let\newline\\\arraybackslash\hspace{0pt}}m{\columnwidth} }
\centering
\begin{tikzpicture}
\begin{axis}[axis lines=left,domain=0:70,axis equal image,
  width=\textwidth*0.55,
  restrict y to domain=0:70, every tick label/.append style={font=\scriptsize},
  xlabel style={align=center},
  xlabel={\scriptsize 3D annotation time [min] \\ (a)}, 
  ylabel={\scriptsize 2D annotation time - 3 MIPs [min]},
  legend style={
    font=\scriptsize,
    draw=none},
  legend pos=outer north east]
  \addplot[only marks,red,mark=o]table[x=3D,y=2D]{user_study/results.dat};
  \addplot[] plot (\x,{\x});
  \addplot[] plot (\x,{0.5*\x});
  \node[rotate=45] at (axis cs:62,59) {\scriptsize equal time};
  \node[rotate=27] at (axis cs:50,22) {\scriptsize 2D annotation $2\times$ faster};
\end{axis}
\end{tikzpicture}
\quad
\quad
\begin{tikzpicture}
\begin{axis}[axis lines=left,domain=0:70,axis equal image,
  width=\textwidth*0.55,
  restrict y to domain=0:70, every tick label/.append style={font=\scriptsize},
  xlabel style={align=center},
  xlabel={\scriptsize 3D annotation time [min] \\ (b)}, 
  ylabel={\scriptsize 2D annotation time - 2 MIPs [min]},
  legend style={
    font=\scriptsize,
    draw=none},
  legend pos=outer north east]
  \addplot[only marks,red,mark=o]table[x=3D,y=2D-2MIP]{user_study/results.dat};
  \addplot[] plot (\x,{\x});
  \addplot[] plot (\x,{0.5*\x});
  \node[rotate=45] at (axis cs:62,59) {\scriptsize equal time};
  \node[rotate=27] at (axis cs:50,22) {\scriptsize 2D annotation $2\times$ faster};
\end{axis}
\end{tikzpicture} 
\\

\begin{tikzpicture}
\begin{axis}[axis lines=left,domain=0:70,axis equal image,
  width=\textwidth*0.55,
  restrict y to domain=0:70, every tick label/.append style={font=\scriptsize},
  xlabel style={align=center},
  xlabel={\scriptsize 3D annotation time [min] \\ (c)}, 
  ylabel={\scriptsize 2D annotation time - 1 MIP [min]},
  legend style={
    font=\scriptsize,
    draw=none},
  legend pos=outer north east]
  \addplot[only marks,red,mark=o]table[x=3D,y=2D-1MIP]{user_study/results.dat};
  \addplot[] plot (\x,{\x});
  \addplot[] plot (\x,{0.5*\x});
  \node[rotate=45] at (axis cs:62,59) {\scriptsize equal time};
  \node[rotate=27] at (axis cs:50,22) {\scriptsize 2D annotation $2\times$ faster};
\end{axis}
\end{tikzpicture}
%\end{tabular}
\caption{
%\color{blue}
Annotation times captured during the user study. The volumes of the mouse brain dataset were annotated both in 3D and in 2D by different users to ensure that the users are not familiar with the stack they were annotating. A pair of annotation times is represented as a single point in each of the plots. Plot (a) presents the time needed to annotate 3 MIPs in 2D, the time needed to annotate 2 MIPs is presented in plot (b), and plot (c) depicts the amount of time necessary to annotate 1 MIP for each training volume. 
\label{fig:user}
}
\end{figure*}

%% file: consistency_plots.tex
\begin{figure}
\center
\begin{tikzpicture}[scale=0.6]
  \begin{axis}[ybar,enlargelimits=.12,bar width=5pt,
               xlabel style={font=\large},
               xlabel={distance $d$},
               ylabel style={align=center,font=\large},
               ylabel={\% positive annotations},
               legend style = {
                 at={(0.0,1.05)},
                 anchor=north west
               },
               title style={align=center},
              ] 
    \addplot [draw=black, fill=blue] table [x index=0, y index=1] {consistency_2D_3D.txt}; 
    \addplot [draw=blue, fill=blue!50] table [x index=0, y index=4] {consistency_2D_3D.txt}; 
    \legend{3D annotations,2D annotations}
  \end{axis} 
\end{tikzpicture}
\quad\quad
\begin{tikzpicture} [scale=0.6]
  \begin{axis}[ybar,enlargelimits=.1,bar width=7pt,
               xlabel style={font=\large},
               xlabel={\large distance},
               ylabel style={font=\large},
               ylabel={\% positive 2D labels}] 
    \addplot [draw=blue, fill=blue!50] table [x index=0, y index=1] {consistency_mip_v5.2.txt}; 
  \end{axis} 
\end{tikzpicture}
\caption{
%\color{blue}
\emph{Left}: Consistency of the 2D and 3D annotations produced in our user study. The bars show the percentage of positive 2D labels within a distance $d$ to the closest projection of a positive 3D label, and the percentage of projections of 3D labels within a distance of $d$ to the closest 2D label, as a function of $d$. 95\% of positive annotations of each type have a corresponding positive annotation of the other type within a distance of less than three pixels, indicating the generally high consistency between the 2D annotations and the 3D ones.
\label{fig:consistency_2D3D}
\emph{Right}: An estimate of the percentage of 2D projection annotations inconsistent across different projections of the same volume. The bars represent the fraction of 2D annotations that have a corresponding annotation in another view at a distance of at most $d$ pixels, as a function of the distance $d$. 20\% of the annotations appear to be inconsistent, but almost never by more than three pixels.
\label{fig:consistency_mip}
}
\end{figure}
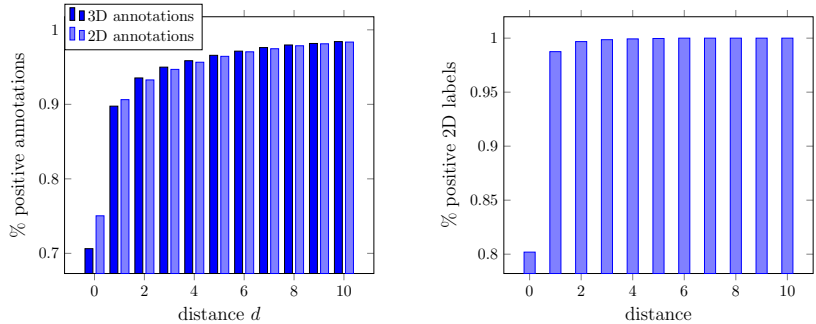

%% file: consistency.tex
\begin{figure}
\center
\begin{tabular}{c@{\hspace{2cm}}c}
\begin{tikzpicture}[scale=0.8]
\pgfmathsetmacro{\cubesize}{2}
\pgfmathsetmacro{\cubeoffset}{1}
\pgfmathsetmacro{\rad}{0.5}
\pgfmathsetmacro{\projoffset}{.5}
%\node[] at (0,-\cubeoffset-\projoffset-0.5,\cubeoffset) (b) {\large 3D image};

%\filldraw[fill=gray!50,draw=gray] (0.5*\rad,0.5*\rad,0.5*\rad) -- ++(-\rad,0,0) -- ++(0,-\rad,0) -- ++(\rad,0,0) -- cycle;
%\filldraw[fill=gray!50,draw=gray] (0.5*\rad,0.5*\rad,0.5*\rad) -- ++(-\rad,0,0) -- ++(0,0,-\rad) -- ++(\rad,0,0) -- cycle;
%\filldraw[fill=gray!50,draw=gray] (0.5*\rad,0.5*\rad,0.5*\rad) -- ++(0,0,-\rad) -- ++(0,-\rad,0) -- ++(0,0,\rad) -- cycle;

\draw[gray] (\cubeoffset,\cubeoffset,\cubeoffset) -- ++(-\cubesize,0,0) -- ++(0,-\cubesize,0) -- ++(\cubesize,0,0); % -- cycle;
\draw[gray] (\cubeoffset,\cubeoffset,\cubeoffset) -- ++(-\cubesize,0,0) -- ++(0,0,-\cubesize) -- ++(\cubesize,0,0); % -- cycle;
\draw[gray] (\cubeoffset,\cubeoffset,\cubeoffset) -- ++(0,0,-\cubesize) -- ++(0,-\cubesize,0) -- ++(0,0,\cubesize) -- cycle;

\draw[gray] (\cubeoffset,\cubeoffset+1,\cubeoffset) -- ++(-\cubesize,0,0) -- ++(0,0,-\cubesize) -- ++(\cubesize,0,0) -- cycle;
\draw[gray] (\cubeoffset+1,\cubeoffset,\cubeoffset) -- ++(0,0,-\cubesize) -- ++(0,-\cubesize,0) -- ++(0,0,\cubesize) -- cycle;

\filldraw[draw=red,thick,fill=red!50] (-0.5*\rad,\cubeoffset+1,-0.5*\rad) -- ++(\rad,0,0) -- ++(0,0,\rad) -- ++(-\rad,0,0) -- cycle;
\draw[draw=red,dashed] (-0.5*\rad,\cubeoffset+1,+0.5*\rad) -- ++(0,-2-0.5*\rad,0) -- ++(\rad,0,0) -- ++(0,2+0.5*\rad,0); % -- cycle;
\draw[draw=red,dashed] (0.5*\rad,\cubeoffset+1,-0.5*\rad) -- ++(0,-2-0.5*\rad,0) -- ++(0,0,\rad) -- ++(0,2+0.5*\rad,0); % -- cycle;
\filldraw[draw=blue,thick,fill=blue!50] (\cubeoffset+1,-0.5*\rad,-0.5*\rad) -- ++(0,0,\rad) -- ++(0,\rad,0) -- ++(0,0,-\rad) -- cycle;
\draw[draw=blue,dashed](\cubeoffset+1,-0.5*\rad,+0.5*\rad) -- ++(-2-0.5*\rad,0,0) -- ++(0,\rad,0) -- ++(2+0.5*\rad,0,0); % -- cycle;
\draw[draw=blue,dashed](\cubeoffset+1,+0.5*\rad,-0.5*\rad) -- ++(-2-0.5*\rad,0,0) -- ++(0,0,\rad) -- ++(2+0.5*\rad,0,0); % -- cycle;

\draw[->] (\cubeoffset+1,\cubeoffset+1,\cubeoffset) -- ++(0,0,-\cubesize);
\node[] at (\cubeoffset+1+0.0,\cubeoffset+1-0.2,\cubeoffset-\cubesize) {$z$};
\pgfmathsetmacro{\tksz}{0.2}
\draw[dotted] (0,\cubeoffset+1,+0.0*\rad) -- ++(\cubeoffset+1+\tksz,0,0);
\draw[dotted] (\cubeoffset+1,0,-0.0*\rad) -- ++(0,\cubeoffset+1+\tksz,0);
\end{tikzpicture}
&
% inconsistent
\begin{tikzpicture}[scale=0.8]
\pgfmathsetmacro{\cubesize}{2}
\pgfmathsetmacro{\cubeoffset}{1}
\pgfmathsetmacro{\rad}{0.5}
\pgfmathsetmacro{\projoffset}{.5}
%\node[] at (0,-\cubeoffset-\projoffset-0.5,\cubeoffset) (b) {\large 3D image};

%\filldraw[fill=gray!50,draw=gray] (0.5*\rad,0.5*\rad,0.5*\rad) -- ++(-\rad,0,0) -- ++(0,-\rad,0) -- ++(\rad,0,0) -- cycle;
%\filldraw[fill=gray!50,draw=gray] (0.5*\rad,0.5*\rad,0.5*\rad) -- ++(-\rad,0,0) -- ++(0,0,-\rad) -- ++(\rad,0,0) -- cycle;
%\filldraw[fill=gray!50,draw=gray] (0.5*\rad,0.5*\rad,0.5*\rad) -- ++(0,0,-\rad) -- ++(0,-\rad,0) -- ++(0,0,\rad) -- cycle;

\draw[gray] (\cubeoffset,\cubeoffset,\cubeoffset) -- ++(-\cubesize,0,0) -- ++(0,-\cubesize,0) -- ++(\cubesize,0,0); % -- cycle;
\draw[gray] (\cubeoffset,\cubeoffset,\cubeoffset) -- ++(-\cubesize,0,0) -- ++(0,0,-\cubesize) -- ++(\cubesize,0,0); % -- cycle;
\draw[gray] (\cubeoffset,\cubeoffset,\cubeoffset) -- ++(0,0,-\cubesize) -- ++(0,-\cubesize,0) -- ++(0,0,\cubesize) -- cycle;

\draw[gray] (\cubeoffset,\cubeoffset+1,\cubeoffset) -- ++(-\cubesize,0,0) -- ++(0,0,-\cubesize) -- ++(\cubesize,0,0) -- cycle;
\draw[gray] (\cubeoffset+1,\cubeoffset,\cubeoffset) -- ++(0,0,-\cubesize) -- ++(0,-\cubesize,0) -- ++(0,0,\cubesize) -- cycle;

\filldraw[draw=red,thick,fill=red!50] (-0.5*\rad,\cubeoffset+1,-0.0*\rad) -- ++(\rad,0,0) -- ++(0,0,\rad) -- ++(-\rad,0,0) -- cycle;
\draw[draw=red,dashed] (-0.5*\rad,\cubeoffset+1,+1.0*\rad) -- ++(0,-2-0.5*\rad,0) -- ++(\rad,0,0) -- ++(0,2+0.5*\rad,0); % -- cycle;
\draw[draw=red,dashed] (+0.5*\rad,\cubeoffset+1,-0.0*\rad) -- ++(0,-2-0.5*\rad,0) -- ++(0,0,\rad) -- ++(0,2+0.5*\rad,0); % -- cycle;
\filldraw[draw=blue,thick,fill=blue!50] (\cubeoffset+1,-0.5*\rad,-1.0*\rad) -- ++(0,0,\rad) -- ++(0,\rad,0) -- ++(0,0,-\rad) -- cycle;
\draw[draw=blue,dashed](\cubeoffset+1,-0.5*\rad,+0.0*\rad) -- ++(-2-0.5*\rad,0,0) -- ++(0,\rad,0) -- ++(2+0.5*\rad,0,0); % -- cycle;
\draw[draw=blue,dashed](\cubeoffset+1,+0.5*\rad,-1.0*\rad) -- ++(-2-0.5*\rad,0,0) -- ++(0,0,\rad) -- ++(2+0.5*\rad,0,0); % -- cycle;

\draw[->] (\cubeoffset+1,\cubeoffset+1,\cubeoffset) -- ++(0,0,-\cubesize);
\node[] at (\cubeoffset+1+0.0,\cubeoffset+1-0.2,\cubeoffset-\cubesize) {$z$};
\pgfmathsetmacro{\tksz}{0.2}
\draw[dotted] (0,\cubeoffset+1,+0.5*\rad) -- ++(\cubeoffset+1+\tksz,0,0);
\draw[dotted] (\cubeoffset+1,0,-0.5*\rad) -- ++(0,\cubeoffset+1+\tksz,0);
\end{tikzpicture}
\\
(a) 
&
(b)
\end{tabular}
\caption{
%\color{blue}
A pair of consistent (a) and inconsistent (b) MIP annotations. A single pixel has been annotated as foreground in each of the two projections (in red and blue). Consistent annotations co-occur along the common dimension ($z$), while the inconsistent annotations do not. For inconsistent annotations, the gradient of our loss function is distributed over a larger number of voxels. The analysis of consistency of annotations performed independently for different projections is presented in section~\ref{sec:study}.
\label{fig:consistency}
}
\end{figure}
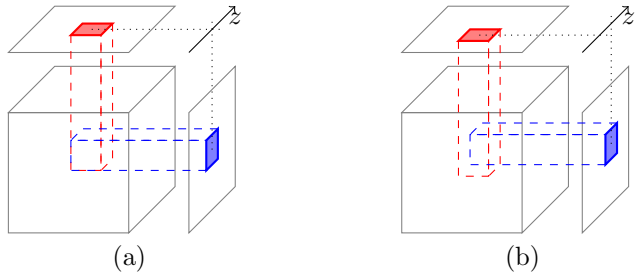

%% file: tab2d3d.tex
% !TEX root = ../main.tex
% !TEX spellcheck = en-US

\begin{table*}
\footnotesize
\newcommand{\fxw}[1]{\makebox[\widthof{16.0}][r]{#1}}
\newcommand{\fxv}[1]{\makebox[\widthof{00.0/00.0}][l]{#1}}
\setlength{\tabcolsep}{4pt}
\centering
\caption{F1 score performance and corresponding time savings. \label{tab:2d3d}}
\begin{tabular}{lccccc}
% & \multicolumn{6}{c}{F1 score [\%] and  diff.\ to NN/3D annot.}& est.\ time gain [\%] \\
% & \multicolumn{1}{c}{axons} & \multicolumn{2}{c}{retina} &  \multicolumn{2}{c}{brain} & \\
&\multicolumn{4}{c}{F1 score}&Time saved\textsuperscript{a} [\%]\\
&Axons&Retina&Angiography&Mouse\\
\hline
 UNet/3D annot.                        &     75.4  &\bf{81.5} &\bf{77.6} &   {80.2}        & \fxw{0}  \\
 UNet/3 MIP per volume              & \bf{78.1} &    78.2  &    75.9  &{\bf{82.2}} & \fxw{35} \\
\hline
 UNet/2 MIP per volume              &      75.0 &     77.8 &     74.8 & 80.0  & \fxw{55} \\
 UNet/1 MIP per volume              &      72.3 &     39.0 &     57.7 & 50.1  & \fxw{70} \\
\hline
%MDOF 
\cite{Turetken13c}   &      58.8 &     77.1 &     22.7 &    18.1       & \fxw{100}\\
%Slice annot.~
\cite{Cicek16}            &      70.8 &     75.8 &     74.1 &    67.5       & \fxw{35}\textsuperscript{b} \\
% Centerline Detection 
\cite{Sironi16a} &      68.5 &     62.6 &     50.3 &    53.6       & \fxw{0}  \\
\hline \\[-2mm]
\multicolumn{6}{l}{\textsuperscript{a} The perc.\ of time saved w.r.t.\ 3D annotation, as estimated in the user study.}\\
\multicolumn{6}{l}{\textsuperscript{b} Slice annotation was assumed to be equally time-consuming as MIP annotation.}\\
%\multicolumn{6}{l}{\textsuperscript{c} Trained on new 2D MIP annotations, as opposed to projections of 3D annotations.}
\end{tabular}
\end{table*}

%% file: conclusion.tex
% !TEX root = main.tex
% !TEX spellcheck = en-US

\section{Conclusion}

We have proposed a method for training DNNs to segment 3D images of linear structures using only annotations of 2D maximum intensity projections of the training data instead of full 3D annotations. We demonstrated that this results in decreased annotation requirements without loss of performance. To this end, we have exploited properties of visual hulls that are not specific to linear structures. In future work, we therefore intend to show that the scope of this technique is in fact much broader, for example by applying it to 3D membrane extraction.

\paragraph*{Acknowledgement} 
We would like to thank Huanxiang Lu, Ying Shi and Felix Sch\"urmann from the Blue Brain Project for sharing the \Mouse{} data. We also thank Ying and Huanxiang for giving us an overview of the practical aspects of neuron delineation. 

MK would like to thank the European Commission for the support received from the FastProof PoC Grant.

AM acknowledges funding from the Swiss National Science Foundation.